\documentclass[a4paper,11pt]{article}
\RequirePackage{amsthm,amsmath,amsfonts,amssymb}
\usepackage[numbers, authoryear]{natbib}
\RequirePackage[colorlinks,citecolor=blue,urlcolor=blue]{hyperref}
\RequirePackage{graphicx}
\usepackage[top=2cm,bottom=2cm,right=2cm,left=2cm]{geometry}
\usepackage{algorithm}
\usepackage{algorithmic}
\usepackage{enumerate}
\usepackage[symbol]{footmisc}

\title{\textbf{Marginal vs Conditional VIM}}
\author{Mohammad Kaviul Anam Khan  \\ \textsc{email:} \textcolor{blue}{\href{mailto:kaviul.khan@mail.utoronto.ca}{kaviul.khan@mail.utoronto.ca}}}
\newtheorem{theorem}{Theorem}

\begin{document}

\begin{center}
	\Large
	Marginal and Conditional Importance Measures from Machine Learning Models and Their Relationship with Conditional Average Treatment Effect
	
	%	\vspace{0.4cm}
	%	\large
	%	Thesis Subtitle
	\small
	\vspace{0.4cm}
	Mohammad Kaviul Anam Khan\footnote[1]{The Hospital for Sick Children. Email: kaviul.khan@mail.utoronto.ca}, Olli Saarela\footnote[2]{Dalla Lana School of Public Health University ot Toronto. Email: olli.saarela@utoronto.ca} and Rafal Kustra\footnote[2]{Dalla Lana School of Public Health University ot Toronto. Email: r.kustra@utoronto.ca} \\
	January 26, 2025
	
	\vspace{0.9cm}
	
\end{center}

\begin{abstract}
Interpreting black-box machine learning models is challenging due to their strong dependence on data and inherently non-parametric nature. This paper reintroduces the concept of importance through ``Marginal Variable Importance Metric" (MVIM), a model-agnostic measure of predictor importance based on the true conditional expectation function. MVIM evaluates predictors' influence on continuous or discrete outcomes. A permutation-based estimation approach, inspired by \citet{breiman2001random} and \citet{fisher2019all}, is proposed to estimate MVIM. MVIM estimator is biased when predictors are highly correlated, as black-box models struggle to extrapolate in low-probability regions. To address this, we investigated the bias-variance decomposition of MVIM to understand the source and pattern of the bias under high correlation. A Conditional Variable Importance Metric (CVIM), adapted from \citet{strobl2008conditional}, is introduced to reduce this bias. Both MVIM and CVIM exhibit a quadratic relationship with the conditional average treatment effect (CATE). 
\end{abstract}

%\begin{keyword}
%\kwd{Variable Importance}
%\kwd{Machine Learning Models}
%\kwd{Conditional Average Treatment Effects}
%\end{keyword}
%
%\end{frontmatter}

	\section{Introduction}
	
	Using modern machine learning techniques, health outcomes for patients can be accurately predicted or classified \citep{howell2012symptom,hawken2010utility,fisher2019machine,cruz2006applications,molnar2020interpretable}. For example, an enhanced number of patient characteristics can be used to successfully predict the survival time of cancer patients using artificial neural networks (ANN) \citep{liestbl1994survival,ripley2001neural}. Studies have also used black-box machine learning models to analyze genetic data \citep{libbrecht2015machine}. However, their applicability in the various scientific fields can be hampered due to the lack of interpretability of the predictors relationship with the outcome from the models. Often, in clinical or health science research, parametric statistical models are preferred over black-box models because they are easily interpretable. Although some work has been published to interpret outputs from machine learning methods, they are mostly data driven and do not have any statistical or causal interpretation, which limits their use in many scientific domains, e.g., in clinical and public health. \\
	
	In the last few decades, some developments have been made to explain the output of black-box machine learning models \citep{molnar2020interpretable}. The popular methods include the Variable Importance Metric (VIM; \citealp{breiman2001random}), LIME (Local Interpretable Model-agnostic Explanations; \citealp{ribeiro2016should}), SHAP (SHapley Additive exPlanations; \citealp{lundberg2017unified}) and leave one covariate out (LOCO) based estimators \citep{rinaldo2019bootstrapping,verdinelli2024feature}. For example, in a recent study \citep{weller2021predicting} the authors were able to find risk factors for suicidal thoughts and behavior using multiple machine learning models using SHAP. However, it is not yet clear whether these methods can quantify a causal relationship between predictors and outcomes.\\
	
	\citet{breiman2001random} proposed the VIM for random forests to identify important variables for prediction. The VIM provides the ranking of the predictors based on changes in the prediction error. The VIM approach is widely applicable and can be used effectively for conventional linear models, generalized linear models, and generalized additive models \citep{gregorutti2017correlation, fisher2019all, hooker2021unrestricted}. \citet{hooker2007generalized} showed the VIM equates to the Sobol indices \citep{fel2021look} when all predictors are independent. Numerous studies, however, have found that the permutation-based importance of random forests can produce highly misleading diagnostics, particularly when there is a strong dependence between features / predictors \citep{strobl2007bias, strobl2008conditional, hooker2021unrestricted}. Using simulation studies, \citet{hooker2021unrestricted} advocated the use of alternative measures such as conditional permutation-based VIM \citep{strobl2008conditional, debeer2020conditional} to interpret the importance of variables, instead. \citep{chamma2024statistically} showed theoretically and empirically that conditional permutation importance (CPI) overcomes the limitations of standard permutation importance by providing accurate type I error control. \citet{kaneko2022cross} proposed a technique based on cross-validation to calculate VIM that minimizes the effect of predictor correlations using simulations. However, most of these simulation scenarios were built on assuming simple functional relationships between outcomes and predictors. It is unclear, though, how VIM would perform in more complex scenarios, such as nonlinear and nonadditive functional relationships between predictors and outcome in presence of high correlation between the predictors. Furthermore, the question still remains about the use of such an importance measure and whether it is possible to use this measure to discern the causal relationship (if any) between the predictor and the outcome or if the VIM can be represented as a statistical parameter.\\
	
	\citet{diaz2015variable} argued that identifying predictor importance and prediction of an outcome have different goals, and the VIM proposed by \citet{breiman2001random} does not have any clinical or causal relevance. \citet{van2011targeted} created a new definition of variable importance based on a targeted causal parameter of interest, which has been used very effectively for many real-life applications. \citet{fisher2019all}, defined Model Reliance (MR) based on the idea of VIM. \citet{gregorutti2017correlation, hooker2021unrestricted} have also shown similar representations of such a statistical parameter. \citet{fisher2019all} further showed such a metric can be represented as a statistical parameter and also have a causal interpretation. They further proposed techniques for estimating MR and their probabilistic bounds. \citet{fisher2019all}'s  MR proposal was based on some pre-defined models or a model class. It is unclear if this MR or VIM can be represented as a statistical parameter in cases where the true relationship between an outcome and the predictors is completely unknown. \\
	
	One objective of this study is to relate permutation-based and conditional permutation-based importance measures to causal parameters. This work extends existing variable importance methods, while examining their statistical properties across various simulated scenarios, and focuses on complex predictors-outcome relationship and modern black-box model exemplified by popular extension of Gradient Boosting Machine (Friedman cite) implemented in XGboost [cite]. Additionally, it addresses the bias introduced by correlated predictors in permutation-based measures and demonstrates how conditional permutation-based measures can mitigate this bias. The study also explores the relationship between permutation-based importance measures and leave-one-covariate-out (LOCO) estimators. Finally, it illustrates how, in simple scenarios, marginal permutation-based importance can be derived from conditional permutation-based importance measures. \\
	
	The rest of the paper is structured as follows: in Section 2 we provide the definition of Marginal Variable Importance Metric (MVIM) and its relation to causal parameter. In Section 3 we showed the bias-variance decomposition of MVIM along with some simulation results.  In Section 4 we defined the Conditional Variable Importance Metric (CVIM) and showed its relationship with causal parameter. In Section 5 we showed the relationship between MVIM and CVIM for linear and additive models. Section 6 concludes the findings of the paper with discussion.

	\section{Marginal Variable Importance Metric (MVIM)}
	\label{mvim}
	
	The variable importance metric (VIM) developed by \citet{breiman2001random} uses permutations to investigate the importance of a predictor which was specifically developed for Random Forests. Since then, many authors have defined the metric at the population level \citep{fisher2019all, gregorutti2017correlation, rinaldo2019bootstrapping, hooker2021unrestricted}. \citet{fisher2019all} redefined the VIM as ``Model Reliance (MR)'' extending Breiman's VIM to the population level. They defined MR based on a predefined class of prediction functions, such as additive, linear, or non-linear prediction functions. In this work we propose a \emph{Marginal Variable Importance Metric (MVIM)} that resembles the MR proposed by \citet{fisher2019all}, with the important difference being that we define the metric based on the true conditional expectation function \( f_0 \), which is assumed to be unknown and may not belong to any predefined class of functions. The definition of the MVIM is as follows. Let $O = (Y, X, \boldsymbol{Z})$ be random vector including outcome $Y$, a predictor of interest $X$ (e.g., exposure or treatment), and multivariate set of covariates $\boldsymbol{Z}$. Let two independent observations from the density $P_{O}(.)$  be $O^{(a)} = \{Y^{(a)}, X^{(a)}, \boldsymbol{Z}^{(a)}\}$ and $O^{(b)} = \{Y^{(b)}, X^{(b)}, \boldsymbol{Z}^{(b)}\}$. The expected squared error loss for the true conditional expectation $f_0$ using $O^{(a)}$ is defined as
	\begin{equation}
		e_{\text{orig}} = \mathbb{E}_{X,\boldsymbol{Z},Y}\left(Y^{(a)} - f_0( X^{(a)}, \boldsymbol{Z}^{(a)})\right)^{2}
		\label{orig}
	\end{equation}
	To calculate the importance of the treatment $X$,  $X^{(a)}$ is switched with $X^{(b)}$ in $O^{(a)}$ and the squared loss is recalculated. The switched loss function can be written as
	\begin{equation}
		\small
		e_{\text{switch}}= \mathbf{E} \mathbb{E}_{Y \mid X^{(a)},\boldsymbol{Z}^{(a)}}\left(Y^{(a)} - f_0( X^{(b)}, \boldsymbol{Z}^{(a)})\right)^{2}
		\label{switch}
	\end{equation}
	Here, the outer expectation in bold represents \\$\mathbf{E} = \mathbb{E}_{X^{(b)}}\mathbb{E}_{X^{(a)}}\mathbb{E}_{Z^{(a)}\mid X^{(a)}}$, or in other words this bold expectation represents an iterative expectation and the subscripts represent the distribution of the random variables. Then the marginal Marginal Variable Importance Metric (MVIM) can be defined as
	\begin{equation}
		\mathcal{MI}_{X} = e_{\text{switch}} - e_{\text{orig}}
		\label{gvim}
	\end{equation}
	In this work we term this as the marginal importance as the importance is defined by keeping the marginal distribution of $X$ unchanged. This metric is slightly different from the previously developed measures \citep{hooker2007generalized, hooker2021unrestricted} since it is defined using the true conditional expectation function \( f_0 \), for which the functional form is completely unknown and cannot be assumed to lie in a pre-specified model class. This metric can be considered as a special case of the MR defined by \citet{fisher2019all}.
	
	%One long equation:
	%%\begin{eqnarray}
	%%	\mu_{\text{normal}} & = & \mu_{x} \frac{C_{s}}{K_{x}C_{x}+C_{s}}  \nonumber\\
	%%	& = & \mu_{\text{normal}} - Y_{x/s}\bigl(1-H(C_{s})\bigr)(m_{s}+\pi /Y_{p/s})\\
	%%	& = & \mu_{\text{normal}}/Y_{x/s}+ H(C_{s}) (m_{s}+ \pi /Y_{p/s}).\nonumber
	%%\end{eqnarray}
	
	\subsection{Expressing MVIM as a Causal Parameter for multinomial and continuous treatments}
	\label{marg.gvim.cause}
	%The procedure is described as follows:
	%
	%\begin{enumerate} 
	%	\item Let $O^{(a)} = (Y^{(a)}, X^{(a)}, Z^{(a)})$ and $O^{(b)} = (Y^{(b)}, X^{(b)}, Z^{(b)})$ be independent random variables from the same joint distribution.
	%	\item with respective realizations  $o^{(a)} = (y^{(a)}, x^{(a)}, z^{(a)})$ and $o^{(b)} = (y^{(b)}, x^{(b)}, z^{(b)})$.
	%	\item Let for the conditional mean $f_0$ let the loss function for $o^{(a)}$ be $L(f_0, o^{(a)})$ and the expected loss as $e_{\text{orig}}(f) = E_{f_0}(L(f_0, O^{(a)}))$.
	%	\item Switch $x^{(a)}$ with  $x^{(b)}$. Define the expected loss as $e_{\text{switch}}(f_0) = E_{f_0}(L(f_0, y^{(a)}, x^{(b)}, z^{(a)}))$.
	%	\item Define  $GVIM_{X}(f_0) = e_{\text{switch}}(f_0) -  e_{\text{orig}}(f_0)$ as the variable importance of $X$.
	%\end{enumerate}
	
	\citet{fisher2019all} showed that the MR defined using the true conditional expectation $f_0$ can be represented as a quadratic function of \emph{conditional average treatment effect (CATE)} for a binary treatment. In this section we show that a similar relationship exists for a multinomial and a continuous predictor. The advantage of MVIM type measure is that because it is a quadratic function of CATE, even if the treatment effect varies by sub-populations, the average treatment effect (ATE) may reduce to zero, but the MR can still identify the importance of the treatment. Let $Y_{x}$ and $Y_{x^*}$ be the potential outcomes under treatments $X = x$ and $X= x^*$. The squared conditional average treatment effect can be expressed as, $\text{CATE}^{2}(x,x^{*},\boldsymbol{z}) = \left(\mathbb{E}\left(Y_{x}\mid \boldsymbol{Z} = \boldsymbol{z} \right)- \mathbb{E}\left(Y_{x^*}\mid \boldsymbol{Z} = \boldsymbol{z}\right)\right)^{2}$. Let us assume the strong ignorability of the treatment assignment mechanism, which states that, $0 < P(X = x\mid \boldsymbol{Z} = \boldsymbol{z}) < 1; \forall x$ (positivity) and $(Y_{x}, Y_{x^*} )\perp X\mid \boldsymbol{Z}$ (conditional ignorability), for all values of $\boldsymbol{Z}=\boldsymbol{z}$.The following Theorems \ref{stat.proof.3.1} and \ref{stat.proof.3.2} shows the MVIMs relationship with CATE for multinomial and continuous outcomes. The proofs of Theorem \ref{stat.proof.3.1} and \ref{stat.proof.3.2} are provided in the supplementary documents. \\
	\begin{theorem}
		\label{stat.proof.3.1}
		Let the treatment be multinomial with $K$ categories ($X \in \{1, 2, ..., K\}$), with the marginal probabilities for a specific category of $X=x$ is given as $P(X = x) = p_{x}$. With respect to the true conditional expectation $f_{0}$,  the MVIM in Equation \eqref{gvim} can be re-written as
		\begin{equation}
			\mathcal{MI}_{X} = \sum_{x = 1}^{K}\sum_{x^* \neq x}p_{x}p_{x^*}\mathbb{E}_{\boldsymbol{Z}\mid X=x}\left(\left[  \mathbb{E}(Y_x \mid \boldsymbol{Z}) - \mathbb{E}(Y_{x^*}\mid\boldsymbol{Z}) \right]^{2}\right)
			\label{vim.1}
		\end{equation}
	\end{theorem}
	\noindent This shows that the MVIM depends on the product of marginal probabilities of treatment levels $x$ and $x^*$. Under the positivity assumption, if $\mathcal{MI}_{X}  = 0$, then $\mathbb{E}_{\boldsymbol{Z}|X=x}\text{CATE}^{2}(x, x^*, \boldsymbol{z}) = 0$, for all $x$. Thus the treatment is not important for the outcome and causally not connected to the outcome $Y$. If $\mathcal{MI}_{X}  > 0$, then the treatment has causal effect and the effect could potentially be different for different sub-populations since examining squared differences prevents potential cancellation of effects with opposite signs. In the case of heterogeneous treatment effect, this measure can still recognize an important treatment variable \citep{fisher2019all}. The MVIM has causal interpretation with respect to the true conditional expectation $f_0$ under the strong ignorability assumptions.   \\
	%\newpage
	\begin{theorem}
		\label{stat.proof.3.2}
		For a continuous treatment the MVIM in Equation \eqref{gvim} can be rewritten as,
		\begin{equation}
			\mathcal{MI}_{X}  =  \int_{X^*} \Bigg( \int_{X} \mathbb{E}_{\boldsymbol{Z} \mid x} 
			\bigg( \bigg[ \mathbb{E}\left(Y_{x} \mid \boldsymbol{Z}\right) 
			- \mathbb{E}\left( Y_{x^{*}} \mid \boldsymbol{Z}\right) \bigg]^2 \bigg)  dP_X(x) \, \Bigg) dP_X(x^*) \
			\label{VIM.cont}
		\end{equation}
	\end{theorem}
	\noindent As $X$ and $X^*$ have the same marginal CDF $P_X(.)$. Again, for a continuous treatment the MVIM can be expressed as a quadratic function of the conditional average treatment effect. Here the random variables $X$ and $X^*$ represent the observed and counterfactual treatment respectively.
	
	\subsection{MVIM and LOCO}
	Using results from \citep{gregorutti2017correlation} it can be easily shown that when $f_0(X, \boldsymbol{Z}) = f_1(X) + f_2(\boldsymbol{Z})$ is an additive function of predictors then MVIM for $X$ can be expressed as,
	\begin{equation}
		\mathcal{MI}_{X}=  2\mathbb{V}\left[f_1(X)\right]
		\label{vim.ad.mod}
	\end{equation}
	where, $\mathbb{V}(X)$ is the marginal variance of $X$. Leave-One-Covariate-Out (LOCO) estimator of importance is another approach of defining importance of a predictor at the population level \citep{rinaldo2019bootstrapping, lei2018distribution}. \citet{lei2018distribution} proposed a model agnostic estimation technique for LOCO estimator. The LOCO importance for additive model is defined as,
	\begin{eqnarray}
		\psi_{loco}(j) & = \mathbb{E}_{Y, X, \boldsymbol{Z}}\left(f_0\left(X, \boldsymbol{Z}\right) - f_2(\boldsymbol{Z})\right)^2 \nonumber\\
		& = \mathbb{V}(f_1(X))
		\label{loco}
	\end{eqnarray} assuming $X$ and $\boldsymbol{Z}$ are independent and all predictors are standardized. Thus, it is trivial to show that $\mathcal{MI}_{X} = 2\psi_{loco}(j)$ when $f_0$ is an additive function of $X$.

	\subsection{Estimating MVIM}
	To develop a model-agnostic estimation technique for the MVIM, we adopted a methodology similar to those outlined by \citet{breiman2001random} and \citet{fisher2019all}. The process begins by fitting a prediction model and calculating the prediction error. Next, a predictor of interest is randomly shuffled, and the prediction error is recalculated. MVIM is then estimated by subtracting the original prediction error from the error after shuffling the predictor. Assuming a finite dataset with \(n\) independent observations and \(K\) predictors, causal parameters are typically estimated using the entire sample dataset. This approach works well for parametric models or targeted maximum likelihood estimation (TMLE), provided the sample size is sufficiently large. However, it is well-established that in-sample prediction errors, or those based on training sets, tend to underestimate the true error \citep{efron1997improvements, tibshirani1993introduction} a problem which becomes more pronounced with modern, highly flexible and adaptive methods. \citet{rinaldo2019bootstrapping} discussed the trade-off between prediction accuracy and inference, noting that while data splitting enhances inference robustness, it may reduce prediction accuracy. Although bagging \citep{breiman1996bagging} is effective for Random Forests, it can lead to unstable predictions when applied to other machine learning models \citep{buhlmann2002analyzing}. As a result, we opted to use subbagging (subsample aggregation) as described by \citet{buhlmann2002analyzing} instead of Bagging. In the Subbagging procedure, a subset (typically two-thirds of the total sample size) is selected without replacement to serve as the training set. The remaining samples form the validation set, which is used to estimate MVIM. This procedure is repeated multiple times, and the final MVIM is obtained by averaging the MVIM estimates from each split. The following algorithm represents the procedure of estimating MVIM.
	\begin{algorithm}[H]
		\caption{MVIM Estimation}\label{alg.gvim}
		\begin{algorithmic}[1]
			\FOR{B = $1$ to $100$}
			\STATE Obtain a Bootstrap sample from the full dataset
			\FOR{k = $1$ to $10$}
			\STATE Split the dataset into training (67\%) and validation (33\%) sets
			\STATE Fit a ML model to the response variable (change) using the all the predictors training dataset.
			\STATE Predict $\hat{f}$ the change scores $y$ from the validation set 
			\STATE calculate $\hat{e}_{\text{orig}} = \frac{1}{n_{\text{vald}}}\sum_{i=1}^{n_{\text{vald}}}\left(y_i - \hat{f}_i\right)^2$
			\FOR{m in $1$ to $5$}
			\STATE Permute the predictor of interest in the validation set. 
			\STATE Recalculate the predictions after permuting as $\hat{f}_{j^{\prime}}$
			\STATE calculate $\hat{e}_{\text{switch}} = \frac{1}{n_{\text{vald}}}\sum_{i=1}^{n_{\text{vald}}}\left(y_i - \hat{f}_{j^{\prime}, i}\right)^2$
			\STATE Calculate $\mathcal{MI}_{mkB} = \hat{e}_{\text{switch} } - \hat{e}_{\text{orig}}$
			\ENDFOR
			\ENDFOR
			\ENDFOR
			\STATE Estimate $\widehat{\mathcal{MI}}$ by averaging the $\mathcal{MI}_{mkB}$s
		\end{algorithmic}
	\end{algorithm} 
	
	\subsection{Statistical Properties of MVIM Estimator}
	\label{sec.Sim}
	We evaluated the properties of the MVIM estimator through simulations, comparing estimates derived from different training sample sizes to the true MVIM values in the population. The simulations were constructed to reflect the complex scenarios encountered in many scientific fields, such as clinical and public health. For this simulation study we constructed the response $Y$ using the following model inspired by \citet{friedman1991multivariate},
	\begin{equation}
		\begin{split}
			Y  = & 2X_1 - 4X_1C_1 + 2C_1 + 2\log(\mid X_2 X_3\mid + 0.1)  \\
			+ & (X_4 - 0.5)^3 - 2X_5 + 2\sin(\pi U_1U_2)  \\
			- & \mathcal{I}(C_2 = 2) + 2 \mathcal{I}(C_2 = 3)    + \epsilon
		\end{split}
		\label{sim.mars}
	\end{equation}
	The coefficients were selected to obtain a reasonable spectrum of true MVIMs. Here, $\epsilon \sim \text{N}(0,1)$ is the random error. The continuous random predictors $X_2, X_3, X_4, X_5$ were generated from $\text{Normal}(0,1)$ distributions. The categorical predictors  $C_1$ and $C_2$ were generated from binomial $(0.5, 0.5)$ and multinomial $(1/3, 1/3, 1/3)$ distributions respectively, which were also independent. The variables $U_1$ and $U_2$ were drawn independently of the other predictors from a Uniform(-1,1) distribution. Another 45 variables $X_6 - X_{50}$, were generated from $N(0,1)$, which were not related to the outcome, or the other predictors, and thus had no importance. To investigate the effect of predictor correlation on MVIM estimation I developed three scenarios:
	
	\begin{description}
		\item[Scenario 1] All the predictors were considered to be independent and hence $X_1$ was simulated from $\text{Normal}(0,1)$ distribution. This scenario is labeled as the \emph{``completely independent scenario''}.
		\item[Scenario 2] Again, $X_1$ was simulated from $\text{Normal}(0,1)$ distribution. Additionally the correlation between $X_1$ and $X_5$ was fixed at 0.9. This scenario is labeled as the \emph{``simple correlation scenario''}.
		\item[Scenario 3] Here, $X_1$ was generated using the following equation,
		\begin{equation}
				X_1  =  -0.5 + C_1 -0.5X_2 + 0.5X_3 + 0.3X_4  0.3X_5+  \nu; \text{ }\nu \sim N(0,\sigma^2_x = 0.07)
			\label{x.gen}
		\end{equation}
		This scenario is labeled as the \emph{``multivariate correlation scenario''}. This scenario ensures that the correlation between $X_1$ and $\mathbb{E}(X_1\mid X_2, X_3, C_1, X_4, X_5)$ is 0.93. 
	\end{description}
	The Equation \eqref{x.gen} was chosen to make sure that the expectation, $\mathbb{E}(X_1) = 0$, and the marginal variance $\text{Var}(X_1) = 1$, which were the same marginal mean and variance of $X_1$ in scenarios 1 and 2. Our interest is to identify importance of every variable, which may differ from situation where importance or causal connection of a one, or few of, specific covariates is of interest. Using the Equation \eqref{sim.mars} enables us to investigate properties of our method for variety of situations: such as predictors with linear, polynomial, oscillating and multiplicative affect on the outcome $Y$, together with interaction, or effect modification terms present, as between $X_1$ and $C_1$.  We generated training sets of sizes 50, 100, 200, 500, 1000, 5000, 10,000 and 50,000. The aim of this simulation was to investigate how accurately a specific method can estimate MVIM of the predictors $X_1-X_5, C_1, C_2, U_1, U_2$, and whether estimated MVIMs of large number of noise variables remains close to zero. At first to calculate the true MVIM estimates, we generated another dataset of size $n_{\text{pop}} = 1,000,000$. This large dataset was considered as a population, from which the true values of the MVIMs were estimated using Monte Carlo methods\footnote{Some of the MVIMs can be calculated without Monte Carlo methods and some cannot be. To be consistent we calculated all the MVIMs using Monte Carlo method. The Monte Carlo estimates closely approximated the actual values, which could be determined analytically.} with very high accuracy. The true $e_{\text{orig}} = \text{Var}(\epsilon) = 1$ for the dataset and the $e_{\text{switch}}$s varied by the predictors. The true MVIMs for each important predictor are reported in the following Table \ref{tab.1}. 
	\begin{table}[H]
		\centering
		\caption{The True MVIMs for all the Predictors in the simulation for all three scenarios}
		\begin{tabular}{lr}  \hline
			Predictor & MVIM \\ \hline
			$X_1$ & 8.00 \\
			$X_2$ & 7.30 \\
			$X_3$ & 7.30 \\
			$X_4$ & 49.50 \\
			$X_5$ & 8.00 \\
			$C_1$ & 10.00 \\
			$C_2$ & 3.11 \\
			$U_1$ & 3.09 \\
			$U_2$ & 3.09 \\
			$X_6 - X_{50}$ & 0.00 \\
			\hline
		\end{tabular}
		\label{tab.1}
	\end{table}
	
	The MVIMs for all the predictors were the same for all three scenarios. \\
	
	In the next step we generated training sets of various sizes and fitted the following models, 
	\begin{enumerate}[(a)]
		\item The \emph{oracle model}, where the model was fitted with the correct functional forms based on the conditional expectation of the predictors using the seven important predictors. The continuous  predictors were then transformed to the functional forms defined in Equation \eqref{sim.mars}. That is I first performed the following transformations: $W_1 = \log\left(\mid X_2 X_3 \mid + 0.1\right), W_2 = (X_4 - 0.5)^3$ and $W_3 = \sin(\pi U_1 U_2)$ and then obtained Least Squares estimates by fitting the following model:
		\begin{equation*}
			\begin{split}
				&	E(Y\mid X_1, C_1, W_1, W_2, W_3, X_5, C_2) = \beta_0 + \beta_1 X_1 + \beta_{12}X_1C_1 \\
				+ & \beta_3W_1 + \beta_4W_2 + \beta_5X_5 + \beta_6 W_3 + \beta_6\mathcal{I}(C_2 = 2) + \beta_7 \mathcal{I}(C_2 = 3)
			\end{split}
		\end{equation*}
		This model was considered as the \emph{``gold standard''}, since it represented the correct functional form of $f_0(\boldsymbol{X})$ and is unbiased and consistent from standard linear regression theory.
		%	\item A linear model where all the predictors were used as linear terms in the model. None of the interactions were used. This model is to be considered as a wrong model to calculate the MVIMs. When the training size was 50 (that is the number of observations were smaller than the number of predictors), the prediction was made by just the average response.
		%	\item A LASSO model, where again all the predictors were used as linear terms in the model. Again, this can be considered as a wrong model. The shrinkage parameter was calculated based on the large validation data set.
		\item A A \emph{GAM} model with a purely additive structure was fitted using cubic splines in the MGCV \citep{mgcv} package in R. The model parameters were set to their default values, as the choice of splines did not significantly influence the results. For the categorical predictors we used the linear terms without considering any interaction with the other predictors. This model was selected as an example of a \emph{``mis-specified model''}. Results from this model indicates about how the MVIM estimates behave if the prediction model is mis-specified. %Since the model was designed to be additive, neither the nonlinear factors nor the interactions with the categorical predictors were taken into account.
		%	\item A  \emph{Random Forest} model where the number of bootstrap trees were set to be 500 and the number of predictors that were selected randomly for each tree was set to be $\sqrt{P}$, where $P$ is the number of predictors. The MVIM was calculated similarly to other models, i.e., I did not use the original importance metric developed by \citet{breiman2001random}. 
		\item A  \emph{XGBoost} model proposed by \citet{chen2016xgboost} was used as an \emph{``example of a black-box model''}. The number of boosting iterations was chosen to be between 100 trees to 5000 trees, the maximum depth of a tree was set to be between 2-6, and the learning rate was varied between 0.05-0.3. These hyper-parameters were chosen based on results obtained during 200 simulations separately for each training size. The hyper-parameters were then varied based on the training size. For the smaller training sizes the learning rate was chosen to be very small (0.05), with large number of trees (5000). The rate was increased along with decreasing number of trees as the training set increased, since a small learning rate with a large number of additive trees did not improve the results.
	\end{enumerate}
	The estimated $e_{\text{orig}}(\hat{f})$ was calculated from a validation set for a specific model using,
	\begin{equation}
		\hat{e}_{\text{orig}}(\hat{f}) = \dfrac{1}{n_v}\left(\sum_{i=1}^{n_v}y_{i} - \hat{f}(\boldsymbol{X})\right)^2
		\label{e.orig.hat}
	\end{equation}
	where $\boldsymbol{X} = (X_1, X_2, ..., X_{52}, C_1, C_2, U_1, U_2)^{\prime}$ is the predictor vector which included all the predictors (both important and non-important). The $e_{\text{switch}}$ was then estimated by,
	\begin{equation}
		\hat{e}_{\text{switch}}(\hat{f}) = \dfrac{1}{n_v}\left(\sum_{i=1}^{n_v}y_{i} - \hat{f}(\boldsymbol{X}^{\prime}_{(j)})\right)^2
		\label{e.switch.hat}
	\end{equation}
	here, $\boldsymbol{X}^{\prime}_{(j)}$ was the predictor matrix where the $j^{th}$ predictor was permuted in the validation set and $n_v$ was the size of the validation set. When the sample size was large ($>5000$), the permutation was performed once. For smaller sample sizes, the permutation was repeated 10 times, and the switched error was computed as the average of the switched errors across the 10 permutations. \\
	
	We simulated a dataset using the equation \eqref{sim.mars} of varying training sizes. Then the models were fitted using two thirds of the data and then estimated the $\mathcal{MI}_j$ for the $jth$ predictor using the remaining one third of the data. Figures \ref{Fig1} to \ref{Fig3} show the box plot of the MVIMs over the 200 simulations by training sizes for the predictors $X_1$, $C_2$, $X_2$, $X_4$ and $X_5$. Box-plots for the remaining predictors are presented in the Supplementary B document.
	
	\subsection{Results from the simulations}

\begin{figure}[H]
		\centering
		\includegraphics[scale = 0.45]{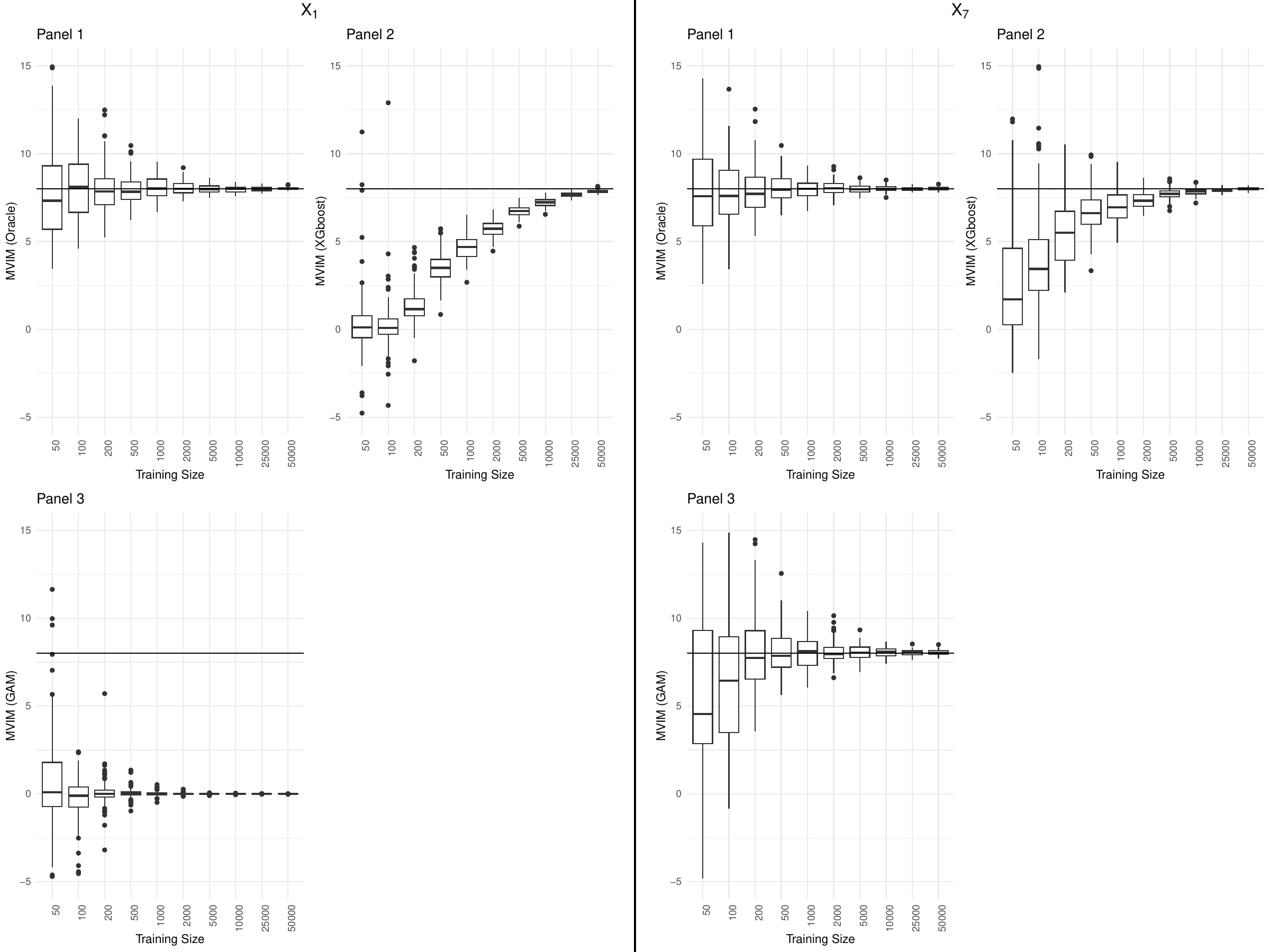}
		\caption{Estimated MVIM for the predictor $X_1$ and $X_5$ for completely independent scenario. Panel 1 shows the Box plots of MVIMs from the oracle model, panel 2 shows the Box plots of MVIM from the XGBoost model and panel 3 shows the Box plots from the GAM model  }
		\label{Fig1}
	\end{figure}

\begin{figure}[H]
		\centering
		\includegraphics[scale = 0.45]{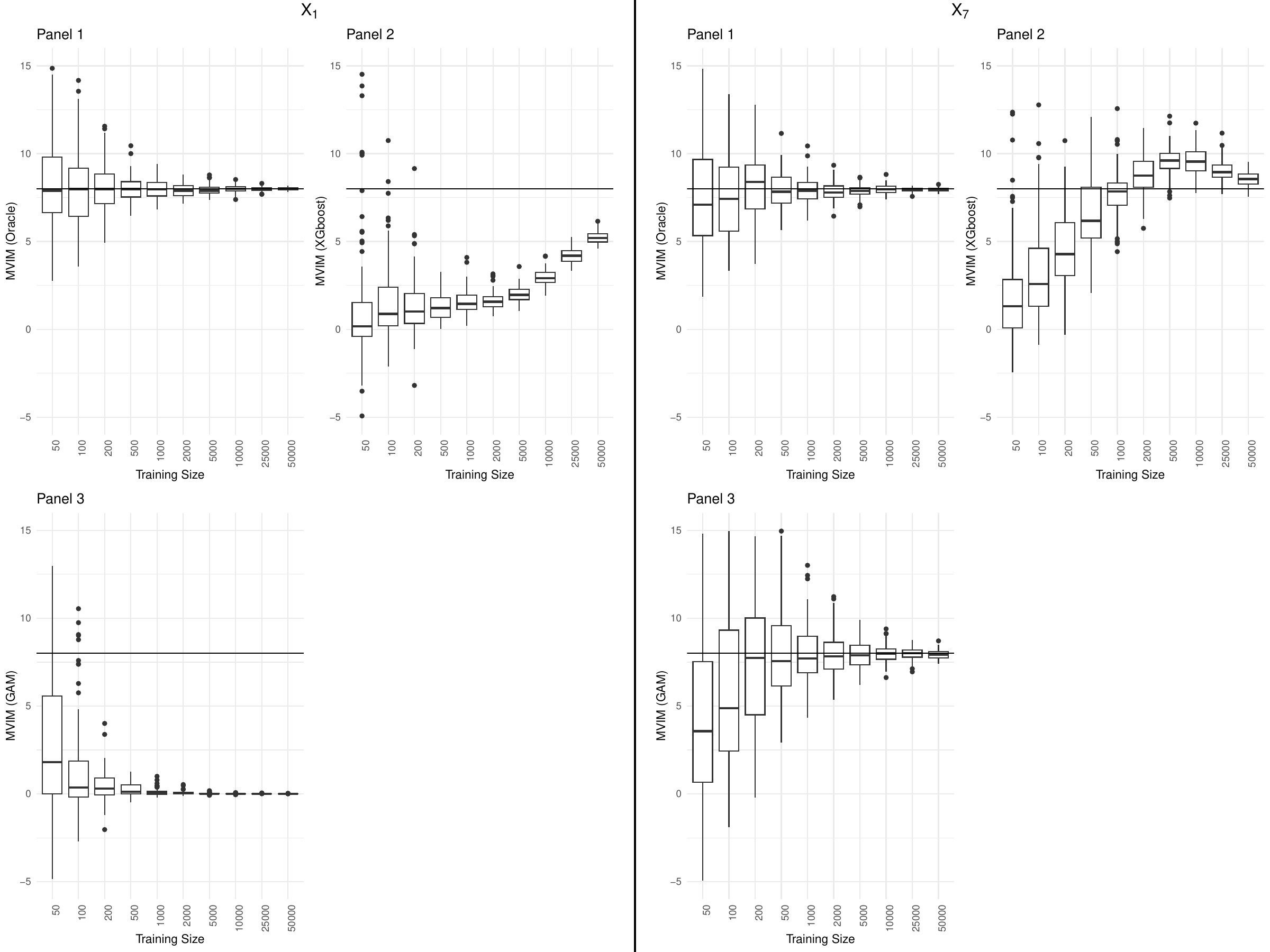}
		\caption{Estimated MVIM for the predictor $X_1$ and $X_5$ for simple correlation scenario. Panel 1 shows the Box plots of MVIMs from the oracle model, panel 2 shows the Box plots of MVIM from the XGBoost model and panel 3 shows the Box plots from the GAM model  }
		\label{Fig2}
	\end{figure}

\begin{figure}[H]
		\centering
		\includegraphics[scale = 0.45]{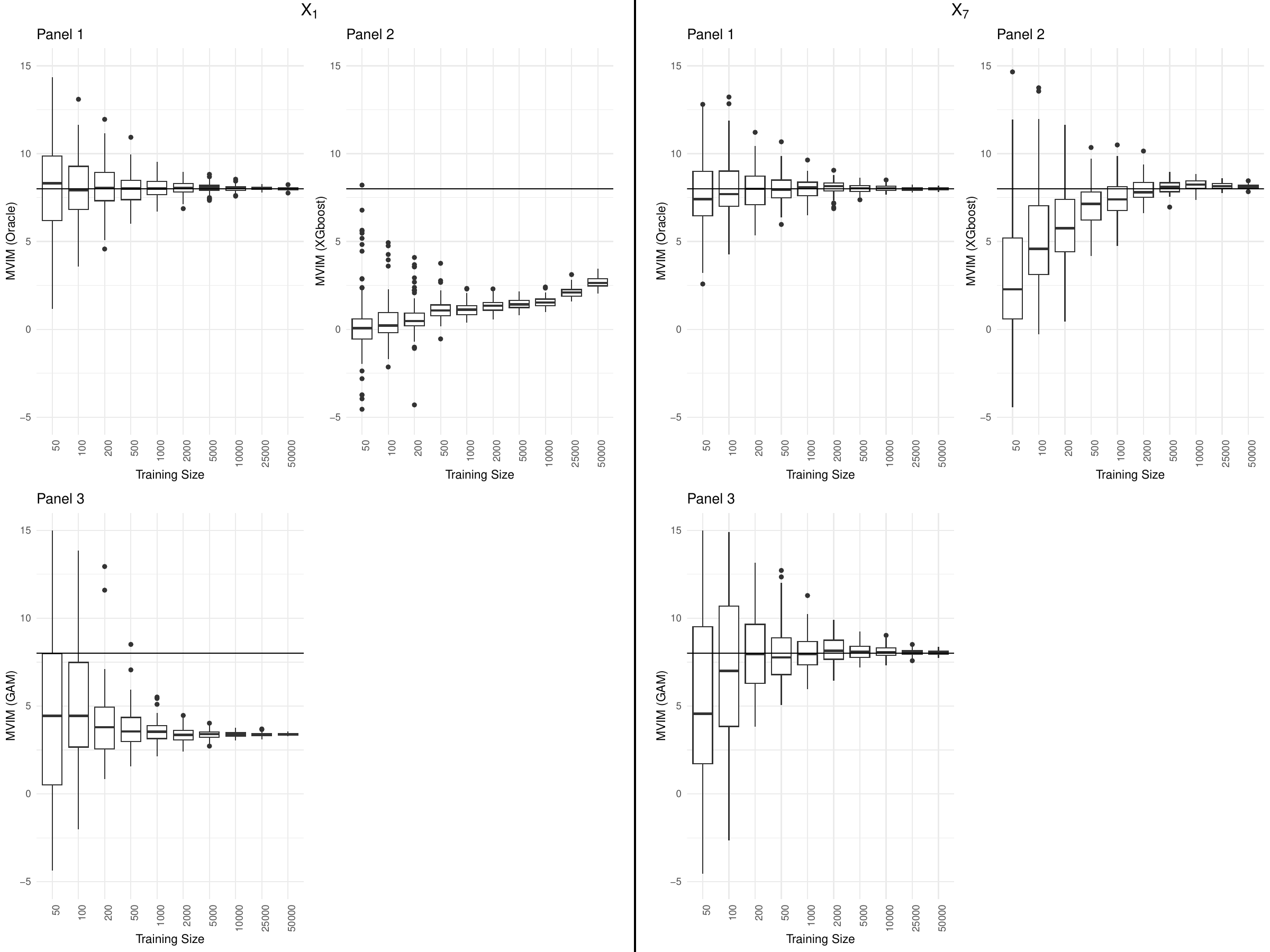}
		\caption{Estimated MVIM for the predictor $X_1$ and $X_5$ for multivariate correlation scenario. Panel 1 shows the Box plots of MVIMs from the oracle model, panel 2 shows the Box plots of MVIM from the XGBoost model and panel 3 shows the Box plots from the GAM model  }
		\label{Fig3}
	\end{figure}

	The box plots in Figures \ref{Fig1} to \ref{Fig3} represent the distributions of the MVIM estimates for the variables $X_1$ and $X_5$. These predictors are related to the outcome and possess different correlation structures as outlined in the three scenarios. In the completely independent scenario presented in Figure \ref{Fig1}, it was observed that the MVIM estimates for both $X_1$ and $X_5$ converged to the true values for the Oracle model (Panel 1) and XGBoost (Panel 2) as training sizes increased. The Oracle model consistently produced unbiased estimates for all training sizes. For XGBoost, the MVIM was initially underestimated, but as the training size increased, the bias in estimating MVIM converged to zero. Since the GAM was mis-specified for $X_1$ (i.e., the interaction between $X_1$ and $C_1$ was not considered), it produced estimates of MVIM close to zero for all training sizes. For $X_5$ the Oracle and the XGBoost models produced similar results as $X_1$ which can be observed in Figure \ref{Fig1}. Similar to the Oracle model the GAM also produced approximately unbiased and consistent estimates of MVIM for $X_5$. This was because $X_5$ had an additive term in the true conditional expectation function, which was within the GAM model class. The MVIM estimates had similar pattern for the rest of the variables as can be observed in the Supplementary B document, i.e., the Oracle model produced unbiased and consistent MVIM estimates, the XGBoost produced asymptotically unbiased and consistent MVIM estimates. The properties of MVIM estimates for a predictor depended on the functional form of the predictor in the conditional expectation function.\\
	
	In the simple correlation scenario, the MVIM estimates for $X_1$ from the Oracle model remained unbiased and consistent, as evidenced in Figure \ref{Fig2}. Conversely, the mis-specified GAM continued to yield estimates close to zero. While XGBoost still provided consistent MVIM estimates, the convergence rate was notably slower compared to the completely independent scenario. For $X_5$ in the simple correlation scenario, estimates remained similar to those in the completely independent scenario when using the Oracle model and GAM (Figure \ref{Fig2}). However, XGBoost initially overestimated MVIMs for training sizes 2000 to 25000, eventually converging to the true MVIM for a training size of 50,000. \\
	
	In the multivariate correlation scenario, the MVIM estimates for $X_1$ followed a similar pattern to the other scenarios when estimated using the Oracle model and GAM, as observed in Figure \ref{Fig3}. However, XGBoost convergence notably slowed even more, again underestimating MVIM, but now for all training sizes. This was likely due to interaction effects between predictors $X_1$ and $C_1$ on the outcome, as well as the small conditional variance of $X_1$ given the five other predictors. The distribution of MVIM estimates for $X_5$, as depicted in Figure \ref{Fig3}, exhibited a very similar pattern to the completely independent scenario across all three models. \\
	
	The box plots for MVIM estimates for the remaining variables for the simple and multivariate correlation scenarios are available in the Supplementary B document. As expected, the Oracle model consistently produced nearly unbiased estimators of MVIMs for all training sizes and variables. Conversely,  GAM yielded biased and inconsistent estimates when the functional forms are mis-specified, though it provided consistent estimates in cases with additive relationships to the outcome. When the training size is very small ($<5000$), MVIMs calculated from XGBoost tend to be underestimated for all variables due to large prediction errors stemming from the small training set size. However, for most predictors, this bias converges to zero with increasing training size. Specifically, biases for variables $X_2$, $X_3$, $X_4$, and $X_5$ all converge to zero with increasing training set sizes. For variables $U_1$ and $U_2$, the means of estimated MVIMs converge to the true MVIM, although there are still some differences between them for a training set size of 50000.
	
%	\newpage
	
%	\begin{figure}[H]
%		\centering
%		\includegraphics[scale = 0.30]{GGBoxplot_simp_17.pdf}
%		\caption{Estimated MVIM for the predictor $X_5$ for completely independent scenario }
%		\label{Fig4}
%	\end{figure}
%	
%	\begin{figure}[H]
%		\centering
%		\includegraphics[scale = 0.30]{GGBoxplot_cor_17.pdf}
%		\caption{Estimated MVIM for the predictor $X_5$ for simple correlation scenario }
%		\label{Fig5}
%	\end{figure}
%	
%	\begin{figure}[H]
%		\centering
%		\includegraphics[scale = 0.30]{GGBoxplot_comp_17.pdf}
%		\caption{Estimated MVIM for the predictor $X_5$ for multivariate correlation scenario }
%		\label{Fig3}
%	\end{figure}
	
	The predictors $X_1$ and $X_5$, both having linear associations with the outcome and similar importance, exhibited different rates of bias reduction for XGBoost. The bias converged much faster to 0 for $X_5$ compared to $X_1$ in the completely independent scenario and the simple correlated scenario, due to $X_1$ having an interaction with $C_1$. In the multivariate correlation scenario, the bias for MVIM estimates of $X_1$ did not completely disappear, even for a training size of 50,000. This was the only case where bias has not converged to 0. Thus, interaction terms and correlations involving multiple predictors can induce bias in estimating MVIM, especially for black-box models like XGBoost. This finding is not surprising, as many other studies \citep{verdinelli2024feature, rinaldo2019bootstrapping, strobl2008conditional, hooker2021unrestricted} have also reported biased estimations of permutation-based feature importance in the presence of strong dependence between predictors. Several studies, such as \citet{hooker2021unrestricted} and \citet{strobl2007bias}, found that the VIM was overestimated by Random Forest; however, their simulations relied on linear prediction functions with different levels of correlations between two or more predictors. \citet{verdinelli2024feature} studied LOCO estimators and showed that the variable importance based on LOCO from random forests can have downward bias. In the simulations, we found both downward and upward bias while estimating MVIM using XGBoost. This can be attributed to a black-box model's inability to extrapolate in low-density regions of the joint distribution of predictors when there is dependence among multiple predictors. As seen in Figures \ref{Fig1} - \ref{Fig3} simple, but strong, correlation of $X_1$ with a single co-predictor ($X_5$) also slowed the converges rates, but substantially less than what was observed in Scenario 3  with correlation structure involving multiple co-predictors. This indicates that correlation distortion can occur when small correlations are present involving multiple predictors. In the multivariate correlation scenario most of the variation in $X_1$ was explained by other predictors. In causal terms, this can be considered a near violation of the positivity assumption. In a practical scenario, it is very likely that an exposure of interest is associated with multiple confounders. As the dimension of predictors increases, low-density regions within the predictor space also increase due to small correlations between several predictors. Switching the values of a predictor can induce those low-density regions resulting in black-box models such as XGBoost to produce predictions very similar to the original prediction. \citet{hooker2021unrestricted} and \citet{fisher2019all} proposed some alternative solutions based on conditional permutations. In the next section, we focus on the bias-variance decomposition of MVIM to understand the source of the bias and what are the possible solutions for debiasing.
	
	\section{Bias-Variance Decomposition of $\widehat{\boldsymbol{MI}}$}
	
	The MVIM is estimated by using the original and switched squared prediction errors. In this section we present the bias-variance decomposition of both the prediction errors. Let, $Y\in \mathbb{R}$ be the outcome and $\boldsymbol{X} = (X_1, X_2, \dots, X_p)$ be the predictors. $\boldsymbol{X}^{\prime}_{j} = (X_1, X_2, \dots, X^{\prime}_j, \\ \dots,  X_p)$, is the vector where the realization of the predictor$X_j$ is switched with it's independent replicate. The true conditional expectation is given by $\mathbb{E}(Y \mid \boldsymbol{X}) = f_0(\boldsymbol{X})$. Prediction model $\hat{f}$ is fitted to the training data. Recall that the bias variance decomposition of the prediction error is expressed as
	\begin{equation}
		\begin{split}
		 \mathbb{E}_{\boldsymbol{X},Y}\left(\hat{e}_{\text{orig}}\right) & = \mathbb{E}_{\boldsymbol{X},Y}\left(Y - \hat{f}(\boldsymbol{X})\right)^2 \\
			& =  \mathbb{E}_{\boldsymbol{X},Y}\left(Y - f_0(\boldsymbol{X})\right)^2 +  	\mathbb{E}_{\boldsymbol{X}}\left(\mathbb{E}_{\mathcal{T}}(\hat{f}(\boldsymbol{X})) -  f_0(\boldsymbol{X})\right)^2    + \mathbb{E}_{\boldsymbol{X}}\mathbb{E}_{\mathcal{T}}\left(\hat{f}({\boldsymbol{X}}) - \mathbb{E}_{\mathcal{T}}(\hat{f}(\boldsymbol{X})) \right)^2 \\
			& =  e_{\text{orig}} +  	\underbrace{\mathbb{E}_{\boldsymbol{X}}\left( \mathbb{E}_{\mathcal{T}}(\hat{f}(\boldsymbol{X})) -  f_0(\boldsymbol{X})\right)^2}_{\text{Bias}^2 \text{(orig)}}  + \underbrace{\mathbb{E}_{\boldsymbol{X}}\left(\hat{f}({\boldsymbol{X}}) -  \mathbb{E}_{\mathcal{T}}(\hat{f}(\boldsymbol{X})) \right)^2}_{\text{Variance (orig)}} 
		\end{split}
		\label{BV1}
	\end{equation}
	Here $ \mathbb{E}_{\mathcal{T}}(.)$ represents the expectation over all training sets. The $e_{\text{orig}}$ is the irreducible error or the true conditional variance. The $\hat{e}_{\text{orig}}$ is the prediction error calculated from a validation set. Similarly the expected value of the switched prediction error can be represented as
	\begin{equation}
		\begin{split}
			\mathbb{E}_{X_j}\mathbb{E} _{\boldsymbol{X}_{-j}}\mathbb{E}_{Y\mid \boldsymbol{X}}\left(\hat{e}_{\text{switch}}\right) & =   \mathbb{E}_{\boldsymbol{X},Y}\left(Y - \hat{f}(\boldsymbol{X}^{\prime}_j)\right)^2 \\
			& =  \mathbb{E}_{\boldsymbol{X},Y}\left(Y - f_0(\boldsymbol{X}^{\prime}_j)\right)^2 + \mathbb{E}_{\boldsymbol{X}}\left(\mathbb{E}_{\mathcal{T}}(\hat{f}(\boldsymbol{X}^{\prime}_j)) -  f_0(\boldsymbol{X}^{\prime}_j)\right)^2  \\
			+  &	\mathbb{E}_{\boldsymbol{X}}\mathbb{E}_{\mathcal{T}}\left(\hat{f}(\boldsymbol{X}^{\prime}_j) - \mathbb{E}_{\mathcal{T}}(\hat{f}(\boldsymbol{X})) \right)^2  \\
			+ & 2\mathbb{E}_{X_j}\mathbb{E}_{ \boldsymbol{X}_{-j}}\left(f_0(\boldsymbol{X} )- f_0(\boldsymbol{X}_{j}^{\prime}))\right)\left(f_0({\boldsymbol{X}_j^{\prime}}) - \mathbb{E}_{\mathcal{T}}(\hat{f}(\boldsymbol{X}_{j}^{\prime})) \right) \\
			= & e_{\text{switch}} +  	\underbrace{\mathbb{E}_{X_j }\mathbb{E}_{ \boldsymbol{X}_{-j}}\left(\mathbb{E}_{\mathcal{T}}(\hat{f}(\boldsymbol{X}_{j}^{\prime})) -  f_0(\boldsymbol{X}_{j}^{\prime})\right)^2}_{\text{Bias}^2 \text{(switch)}} \\
			& + \underbrace{\mathbb{E}_{X_j}\mathbb{E}_{ \boldsymbol{X}_{-j}}\mathbb{E}_{\mathcal{T}}\left(\hat{f}({\boldsymbol{X}_j^{\prime}}) - \mathbb{E}_{\mathcal{T}}(\hat{f}(\boldsymbol{X}_{j}^{\prime})) \right)^2}_{\text{Variance (switch)}}  \\
			& +\underbrace{2\mathbb{E}_{X_j}\mathbb{E}_{ \boldsymbol{X}_{-j}}\left(f_0(\boldsymbol{X}_{j}^{\prime}) - f_0(\boldsymbol{X} )\right)\left(f_0({\boldsymbol{X}_j^{\prime}}) - \mathbb{E}_{\mathcal{T}}(\hat{f}(\boldsymbol{X}_{j}^{\prime}) \right)}_{\text{additional bias term }\delta_j}
		\end{split}
		\label{BV2}
	\end{equation}
	Here, $\boldsymbol{X}^{\prime}_{j}$ represents the predictor vector where the j\textsuperscript{th} predictor $X_j$ is permuted and $\boldsymbol{X}_{-j} = (X_1, X_2, \dots, X_{j-1}, \\ X_{j+1}, \dots, X_P)$, excludes $X_j$ from $\boldsymbol{X}$. Thus, the expected value of the estimated MVIM can be represented as,
	
	\begin{equation}
		\begin{split}
			\mathbb{E}(\widehat{\mathcal{MI}}_j) & = \mathbb{E}_{X_j}\mathbb{E} _{\boldsymbol{X}_{-j}}\mathbb{E}_{Y\mid \boldsymbol{X}}\left(\hat{e}_{\text{switch}}\right) - \mathbb{E}_{\boldsymbol{X},Y}\left(\hat{e}_{\text{orig}}\right) \\
			& =  e_{\text{switch}}  - e_{\text{orig}} + \text{Bias}^2 \text{(switch)} - \text{Bias}^2 \text{(orig)}+  \text{Variance (switch)} - \text{Variance (orig)} + \delta_j  \\
			 &\approx  \mathcal{MI}_j + \delta_j
		\end{split}
		\label{BV3}
	\end{equation}
	For the simplicity of exposition in this section, we assume that $ \text{Bias}^2 \text{(switch)} \approx \text{Bias}^2 \text{(orig)}$ and $\text{Variance (switch)}\\ \approx \text{Variance (orig)}$, which is plausible but may not hold in every scenario. If the value of $\delta_j$ is large then the estimated MVIM will be biased, even if $ \text{Bias}^2 \text{(switch)} \approx \text{Bias}^2 \text{(orig)}$ and $\text{Variance (switch)} \approx \text{Variance (orig)}$. The bias of estimated MVIM can be corrected when $\delta_j$ is added to the estimated MVIM for the predictor $X_j$. The accuracy in estimation of $\delta_j$ depends on the accuracy of estimating $f_0({\boldsymbol{X}_j^{\prime}})$ and $ f_0(\boldsymbol{X})$.  A high absolute value of the additional term $\delta_j$ can induce bias in MVIM estimation even if the assumptions regarding the equity of the bias and variance terms are true. To demonstrate the influence of $\delta_j$ on the MVIM estimation, we calculated the $\delta_j$s for the previously defined simulation scenarios in the section \ref{sec.Sim}.
	
	\subsection{Bias Variance Terms from Completely Independent Scenario}
\begin{figure}[H]
		\centering
		\includegraphics[scale = 0.35]{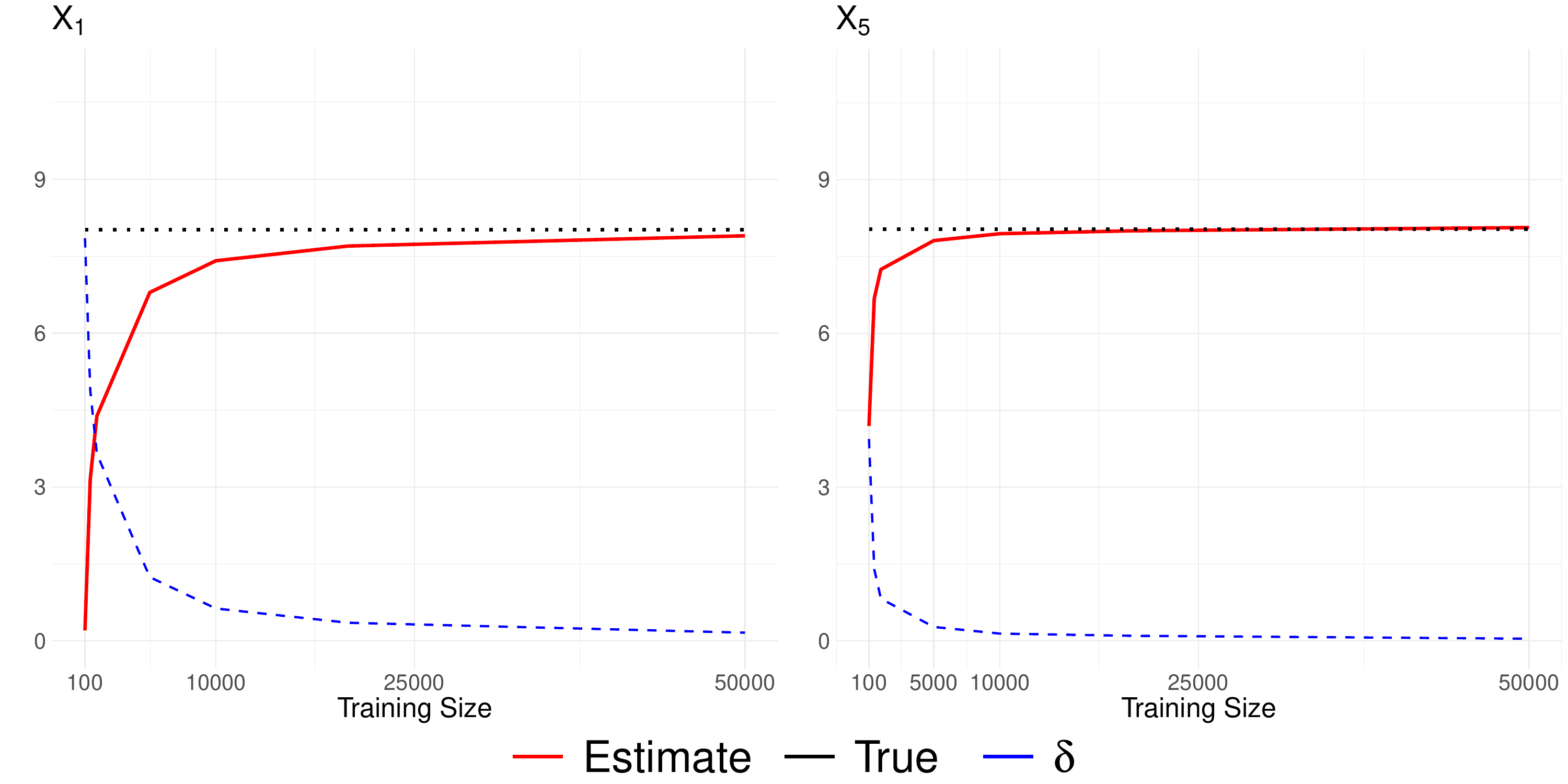}
		\caption{The trajectory of $\delta_j$s and $\widehat{\mathcal{MI}}_j$s of $X_1$ and $X_5$ with increasing training size obtained from the completely independent scenario.}
		\label{Fig4.1}
	\end{figure}

	The terms of the bias and variance components derived in equation \eqref{BV3} for predictors $X_1$, $X_5$, and $X_6$ are reported in Table \ref{tab4.1} in the Appendix section. Here, $X_6$ was a nuisance predictor that had no association with the outcomes or any other predictors. In the completely independent scenario shown in Table \ref{tab4.1}, the assumptions that $\text{Bias}^2 \text{(switch)} \approx \text{Bias}^2 \text{(orig)}$ and $\text{Variance (switch)} \approx \text{Variance (orig)}$ held true. For small training sizes, both the bias squares and the variances were large, and XGBoost struggled to estimate MVIMs accurately for all three predictors. The $\delta_1$ and $\delta_5$ were also sufficiently large for small training sizes. The differences in bias squares and variances, along with $\delta_j$ values, decreased with increasing training sizes for $X_1$ and $X_5$. Eventually, the $\widehat{\mathcal{MI}}$s converged to the true MVIM for both predictors. Figure \ref{Fig4.1} showed the trajectory of the $\delta_j$ values and the MVIM estimates over increasing training sizes. As training size increased, both $\delta_1$ and $\delta_5$ approached 0, and the estimated MVIM approached the true value of MVIM for both predictors. The convergence was faster for $X_5$ compared to $X_1$. Although $X_1$ and $X_5$ were both linearly related to the outcome and had the same true value of MVIM (8.00), $X_1$'s interaction with $C_1$ caused larger values of $\delta_1$ compared to $\delta_5$ for smaller training sizes. For the predictor $X_6$, the $\delta_6$ was always 0, and the estimated MVIMs were approximately 0 for all training sizes.
	
	\begin{table}[H]
		\centering
		\caption{Components of bias-variance decomposition of $\widehat{\mathcal{MI}}$, as described in the equation \eqref{BV3}. The results are obtained for predictors $X_1$, $X_5$ and $X_6$ for the completely independent scenario. The column $\mathcal{MI}_{c}$ was calculated using the equation $\mathcal{MI}_{jc} = \widehat{\mathcal{MI}}_j + \delta_j -  \text{Bias}^2 \text{(switch)} + \text{Bias}^2 \text{(orig)} - \text{Variance (switch)} + \text{Variance (orig)}$. It is expected that the $\mathcal{MI}_{jc}  = \mathcal{MI}$. The slight differences in their values in the table is due to Monte Carlo error.}
		\begin{tabular}{rr|rrrrrrrr} 
			\hline
			& & \multicolumn{2}{c}{$\text{Bias}^2$} & \multicolumn{2}{c}{$\text{Variance}$} & & \multicolumn{3}{c}{$\mathcal{MI}$} \\ \hline
			& $n_\text{train}$ & Switch & Orig & Switch & Orig  & $\delta_j$ & $\widehat{\mathcal{MI}}$ & $\mathcal{MI}_c$ & $\mathcal{MI}$ \\   \hline
			$X_1$ & 100 & 20.09 & 20.05 & 9.93 & 9.94 & 7.85 & 0.20 & 8.01 & 8.00 \\   
			& 500 & 6.52 & 6.49 & 5.35 & 5.35 & 4.88 & 3.16 & 8.01 &  \\   
			& 1000 & 3.66 & 3.64 & 3.99 & 3.99 & 3.65 & 4.39 & 8.01 &  \\   
			& 5000 & 0.71 & 0.70 & 1.92 & 1.91 & 1.24 & 6.79 & 8.01 &  \\  
			& 10000 & 0.31 & 0.30 & 1.32 & 1.30 & 0.63 & 7.41 & 8.02 &  \\  
			& 20000 & 0.15 & 0.15 & 0.89 & 0.86 & 0.35 & 7.70 & 8.02 &  \\  
			& 50000 & 0.11 & 0.11 & 0.42 & 0.39 & 0.16 & 7.90 & 8.03 &  \\  \hline
			$X_5$ & 100 & 20.14 & 20.05 & 9.93 & 9.94 & 3.94 & 4.19 & 8.05 & 8.00 \\   
			& 500 & 6.53 & 6.49 & 5.35 & 5.35 & 1.40 & 6.68 & 8.05 &  \\  
			& 1000 & 3.67 & 3.64 & 3.99 & 3.99 & 0.83 & 7.25 & 8.05 &  \\   
			& 5000 & 0.71 & 0.70 & 1.93 & 1.91 & 0.27 & 7.81 & 8.06 &  \\   
			& 10000 & 0.31 & 0.30 & 1.32 & 1.30 & 0.14 & 7.95 & 8.06 &  \\  
			& 20000 & 0.15 & 0.15 & 0.89 & 0.86 & 0.10 & 8.00 & 8.07 &  \\  
			& 50000 & 0.11 & 0.11 & 0.42 & 0.39 & 0.04 & 8.07 & 8.08 &  \\   \hline
			$X_6$ & 100 & 20.36 & 20.35 & 10.46 & 10.46 & 0.00 & 0.00 & 0.00 & 0.00 \\   
			& 500 & 6.38 & 6.38 & 5.08 & 5.08 & 0.00 & 0.00 & 0.00 &  \\   
			& 1000 & 3.69 & 3.69 & 3.95 & 3.96 & 0.00 & 0.00 & 0.00 &  \\   
			& 5000 & 1.00 & 1.00 & 1.97 & 1.98 & 0.00 & 0.00 & 0.00 &  \\   
			& 10000 & 0.45 & 0.45 & 1.39 & 1.39 & 0.00 & 0.00 & 0.01 &  \\   
			& 20000 & 0.20 & 0.20 & 0.90 & 0.90 & 0.00 & 0.01 & 0.01 &  \\   
			& 50000 & 0.12 & 0.12 & 0.42 & 0.42 & 0.00 & 0.01 & 0.01 &  \\ 
			\hline
		\end{tabular}
		\label{tab4.1}
	\end{table}
	
	\subsection{Bias Variance Terms from Simple Correlation Scenario}
\begin{figure}[H]
		\centering
		\includegraphics[scale = 0.35]{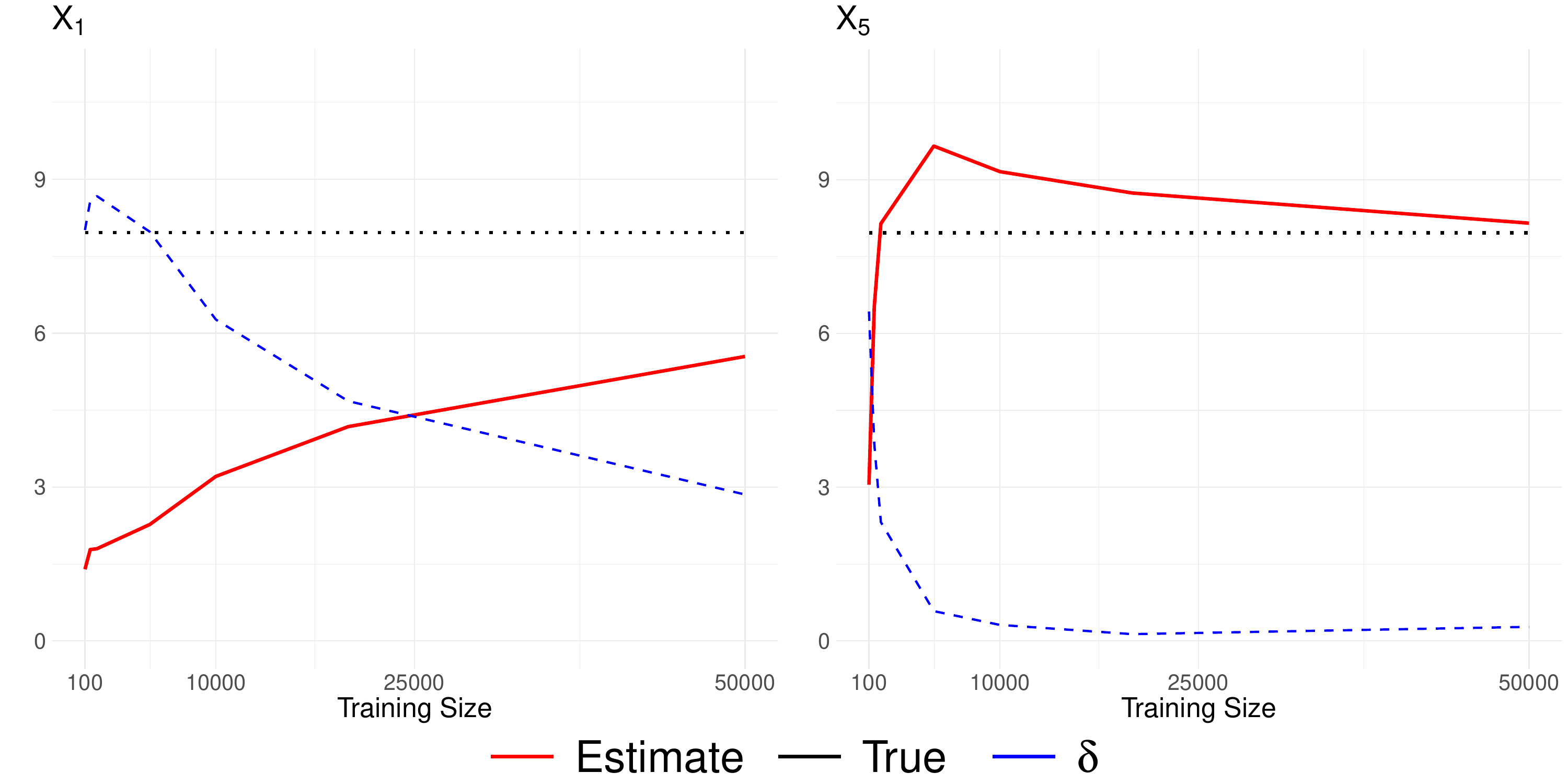}
		\caption{The trajectory of $\delta_1$ and $\widehat{\mathcal{MI}}$ of $X_1$ with increasing training size obtained from the simple correlation scenario}
		\label{Fig4.2}
	\end{figure}

	In the simple correlation scenario, again it can be observed from the Table \ref{tab4.2} that the bias and variance components were large for smaller training sizes and decreased with increasing training size. The values of the bias variance terms for $X_6$ were very similar to the values we observed in Table \ref{tab4.1}. Compared to the completely independent scenario the $\delta_1$ and $\delta_5$ were larger. For all training sizes the $ \text{Bias}^2 \text{(switch)} > \text{Bias}^2 \text{(orig)}$, however the difference is negligible for larger training sizes. For training sizes $\geq 1000$ the MVIM for $X_5$ was overestimated whereas the MVIM for $X_1$ was underestimated for all training sizes as can be seen from the $\widehat{\mathcal{MI}}$ column. From Figure \ref{Fig4.2}, we can see that the $\delta_1$ was decreasing to zero with the increasing training size whereas the estimated $\widehat{\mathcal{MI}}_1$ was approaching to $\mathcal{MI}_1$. However, this convergence rate was much slower compared to the convergence rate for completely independent scenario for both the predictors.  For smaller training sizes the MVIM was overestimated for $X_5$, and converging to the true value with increasing training size. It has been mentioned by \citet{hooker2021unrestricted, strobl2007bias, strobl2008conditional}, that when two predictors are positively correlated with similar importance then their ranking based on VIM obtained from black box models can be higher compared to the rest of the variables. This essentially implies that importance of those predictors tend to be overestimated. This observation was true for $X_5$ in the simulations, however, does not appeared to be true for $X_1$. Here the interaction between $X_1$ and $C_1$ played a role in underestimation of the MVIM for $X_1$.
	
	\begin{table}[H]
		\centering
		\caption{Components of bias-variance decomposition of $\widehat{\mathcal{MI}}$, as described in \eqref{BV3}. The results are obtained for predictors $X_1$, $X_5$ and $X_6$ for the simple correlation scenario. The column $\mathcal{MI}_{c}$ was calculated using the equation $\mathcal{MI}_{jc} = \widehat{\mathcal{MI}}_j + \delta_j -  \text{Bias}^2 \text{(switch)} + \text{Bias}^2 \text{(orig)} - \text{Variance (switch)} + \text{Variance (orig)}$. It is expected that the $\mathcal{MI}_{jc} =\mathcal{MI}$. The slight differences in their values in the table is due to Monte Carlo error.}
		\begin{tabular}{rr|rrrrrrrr} 
			\hline
			& & \multicolumn{2}{c}{$\text{Bias}^2$} & \multicolumn{2}{c}{$\text{Variance}$} & & \multicolumn{3}{c}{$\mathcal{MI}$} \\ \hline
			& $n_\text{train}$ & Switch & Orig & Switch & Orig  & $\delta_j$ & $\widehat{\mathcal{MI}}$ & $\mathcal{MI}_c$ & $\mathcal{MI}$ \\   \hline
			$X_1$ & 100 & 21.63 & 20.35 & 10.63 & 10.46 & 8.01 & 1.40 & 7.96 & 8.00 \\   
			& 500 & 8.48 & 6.38 & 5.39 & 5.08 & 8.58 & 1.78 & 7.96 &  \\   
			& 1000 & 5.97 & 3.69 & 4.19 & 3.96 & 8.67 & 1.80 & 7.96 &  \\   
			& 5000 & 3.05 & 1.00 & 2.21 & 1.98 & 7.98 & 2.27 & 7.96 &  \\   
			& 10000 & 1.75 & 0.45 & 1.60 & 1.39 & 6.27 & 3.21 & 7.97 &  \\   
			& 20000 & 0.95 & 0.20 & 1.04 & 0.90 & 4.68 & 4.18 & 7.98 &  \\   
			& 50000 & 0.42 & 0.12 & 0.53 & 0.42 & 2.85 & 5.55 & 7.99 &  \\   \hline
			$X_5$ & 100 & 21.69 & 20.35 & 10.63 & 10.46 & 6.43 & 3.05 & 7.96 & 8.00 \\   
			& 500 & 8.49 & 6.38 & 5.39 & 5.08 & 3.85 & 6.52 & 7.96 &  \\   
			& 1000 & 5.95 & 3.69 & 4.19 & 3.96 & 2.32 & 8.14 & 7.97 &  \\   
			& 5000 & 3.01 & 1.00 & 2.23 & 1.98 & 0.58 & 9.66 & 7.98 &  \\   
			& 10000 & 1.71 & 0.45 & 1.62 & 1.39 & 0.31 & 9.16 & 7.98 &  \\   
			& 20000 & 0.93 & 0.20 & 1.06 & 0.90 & 0.13 & 8.74 & 7.99 &  \\   
			& 50000 & 0.42 & 0.12 & 0.55 & 0.42 & 0.27 & 8.15 & 8.00 &  \\    \hline
			$X_6$ & 100 & 20.36 & 20.35 & 10.46 & 10.46 & 0.00 & 0.00 & 0.00 & 0.00 \\   
			& 500 & 6.38 & 6.38 & 5.08 & 5.08 & 0.00 & 0.00 & 0.00 &  \\   
			& 1000 & 3.69 & 3.69 & 3.95 & 3.96 & 0.00 & 0.00 & 0.00 &  \\   
			& 5000 & 1.00 & 1.00 & 1.97 & 1.98 & 0.00 & 0.00 & 0.00 &  \\   
			& 10000 & 0.45 & 0.45 & 1.39 & 1.39 & 0.00 & 0.00 & 0.01 &  \\   
			& 20000 & 0.20 & 0.20 & 0.90 & 0.90 & 0.00 & 0.01 & 0.01 &  \\   
			& 50000 & 0.12 & 0.12 & 0.42 & 0.42 & 0.00 & 0.01 & 0.01 &  \\ 
			\hline
		\end{tabular}
		\label{tab4.2}
	\end{table}
	
	\subsection{Multivariate Correlation Scenario}
	\label{mult.cor}
\begin{figure}[H]
		\centering
		\includegraphics[scale = 0.35]{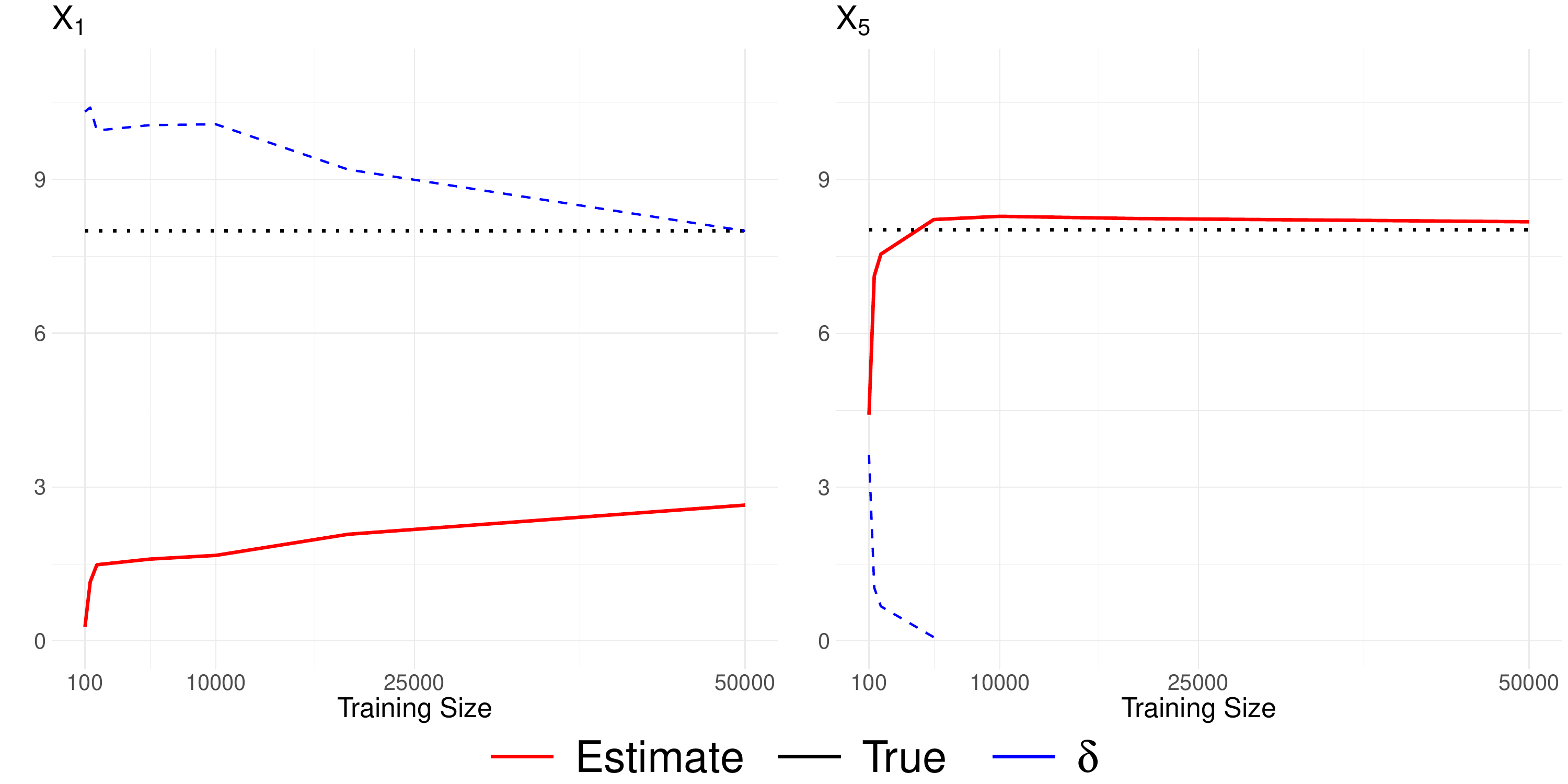}
		\caption{The trajectory of $\delta_1$ and $\widehat{\mathcal{MI}}$ of $X_1$ with increasing training size obtained from the  multivariate correlation scenario scenario}
		\label{Fig4.3}
	\end{figure}

	For this scenario, $X_1$ was weakly to moderately correlated with multiple predictors, and the correlation between $X_1$ and $X_5$ was -0.3, which could be considered a weak correlation. Table \ref{tab4.3} and Figure \ref{Fig4.3} showed that for $X_1$, the $\text{Bias}^2(\text{Switch})$ was higher than the $\text{Bias}^2(\text{Orig})$ for all training sizes. The $\delta_1$ was very large even for a training size of 50,000. In fact, the $\widehat{\mathcal{MI}}$ of $X_1$ was heavily underestimated for all training sizes. From Figure \ref{Fig4.3}, we could see that the $\delta_1$ was larger than the true value of the MVIM and only started to decline for a training size of 5000. The bias and variance components for $X_5$ in Table \ref{tab4.3} were similar to the numbers observed in Table \ref{tab4.1}. The MVIM for $X_5$ was slightly overestimated but to a negligible degree.
	\begin{table}[H]
		\centering
		\caption{Components of bias-variance decomposition of $\widehat{\mathcal{MI}}$, as described in \eqref{BV3}. The results are obtained for predictors $X_1$, $X_5$ and $X_6$ for the simple correlation scenario. The column $\mathcal{MI}_c$ was calculated using the equation $\mathcal{MI}_{jc} = \widehat{\mathcal{MI}}_j + \delta_j -  \text{Bias}^2 \text{(switch)} + \text{Bias}^2 \text{(orig)} - \text{Variance (switch)} + \text{Variance (orig)}$. It is expected that the $\mathcal{MI}_{jc}  = \mathcal{MI}$. The slight differences in their values in the table is due to Monte Carlo error.}
		\begin{tabular}{rr|rrrrrrrr} 
			\hline
			& & \multicolumn{2}{c}{$\text{Bias}^2$} & \multicolumn{2}{c}{$\text{Variance}$} & & \multicolumn{3}{c}{$\mathcal{MI}$} \\ \hline
			& $n_\text{train}$ & Switch & Orig & Switch & Orig  & $\delta_j$ & $\widehat{\mathcal{MI}}$ & $\mathcal{MI}_c$ & $\mathcal{MI}$ \\   \hline
			$X_1$ & 100 & 20.99 & 18.39 & 9.52 & 9.69 & 10.13 & 0.29 & 7.97 & 8.00 \\  
			& 500 & 9.33 & 5.93 & 5.26 & 5.12 & 10.40 & 1.10 & 7.96 &  \\   
			& 1000 & 6.68 & 3.20 & 3.99 & 3.79 & 10.37 & 1.26 & 7.96 &  \\   
			& 5000 & 4.52 & 0.74 & 2.13 & 1.91 & 10.55 & 1.41 & 7.96 &  \\   
			& 10000 & 3.58 & 0.33 & 1.58 & 1.32 & 9.67 & 1.81 & 7.96 &  \\   
			& 20000 & 3.41 & 0.18 & 1.10 & 0.87 & 9.40 & 2.01 & 7.97 &  \\   
			& 50000 & 2.50 & 0.13 & 0.61 & 0.40 & 7.74 & 2.84 & 7.98 &  \\   \hline
			$X_5$ & 100 & 18.41 & 18.39 & 9.69 & 9.69 & 3.61 & 4.45 & 8.03 & 8.00 \\   
			& 500 & 6.05 & 5.93 & 5.13 & 5.12 & 1.04 & 7.12 & 8.03 &  \\   
			& 1000 & 3.36 & 3.20 & 3.80 & 3.79 & 0.71 & 7.50 & 8.04 &  \\   
			& 5000 & 0.98 & 0.74 & 1.94 & 1.91 & 0.12 & 8.19 & 8.04 & \\   
			& 10000 & 0.52 & 0.33 & 1.37 & 1.32 & 0.04 & 8.24 & 8.04 &  \\   
			& 20000 & 0.33 & 0.18 & 0.92 & 0.87 & 0.03 & 8.21 & 8.05 &  \\   
			& 50000 & 0.21 & 0.13 & 0.44 & 0.40 & 0.04 & 8.15 & 8.06 &  \\   \hline
			$X_6$ & 100 & 18.39 & 18.39 & 9.69 & 9.69 & 0.00 & 0.00 & 0.00 & 0.00 \\   
			& 500 & 5.93 & 5.93 & 5.12 & 5.12 & 0.00 & 0.00 & -0.00 &  \\   
			& 1000 & 3.20 & 3.20 & 3.79 & 3.79 & 0.00 & 0.00 & 0.00 &  \\   
			& 5000 & 0.74 & 0.74 & 1.91 & 1.91 & 0.00 & 0.00 & 0.00 &  \\   
			& 10000 & 0.33 & 0.33 & 1.32 & 1.32 & 0.00 & 0.01 & 0.01 &  \\   
			& 20000 & 0.18 & 0.18 & 0.87 & 0.87 & 0.00 & 0.01 & 0.01 &  \\   
			& 50000 & 0.12 & 0.13 & 0.40 & 0.40 & 0.00 & 0.01 & 0.01 &  \\ 
			\hline
		\end{tabular}
		\label{tab4.3}
	\end{table}
	
	\subsection{The Influence of correlation on Estimating MVIM}
	\label{reason.delta}
	
	From Tables \ref{tab4.1} - \ref{tab4.3}, it can be observed that the differences between the variance components (i.e., $\text{Variance (switch)}$ and $\text{Variance (orig)}$) were negligible for all cases. The differences between the bias components (i.e., $\text{Bias}^2 \text{(switch)}$ and $\text{Bias}^2 \text{(orig)}$) were not negligible when high correlation existed between two or more predictors. Recall,
	
	\begin{equation}
		\begin{split}
			&	\text{Bias}^2 \text{(switch)} - \text{Bias}^2 \text{(orig)} \\
			= & \mathbb{E}_{X_j}\mathbb{E}_{\boldsymbol{X}_{-j}}\left(\mathbb{E}_{\mathcal{T}}(\hat{f}(\boldsymbol{X}_{j}^{\prime})) - f_0(\boldsymbol{X}_{j}^{\prime})\right)^2 - \mathbb{E}_{\boldsymbol{X}}\left( \mathbb{E}_{\mathcal{T}}(\hat{f}(\boldsymbol{X})) - f_0(\boldsymbol{X})\right)^2
		\end{split}
		\label{bias.comp}
	\end{equation}
	When high correlation existed between two or more predictors, $\text{Bias}^2 \text{(switch)}$ was greater than $\text{Bias}^2 \text{(orig)}$, implying that the bias in switch prediction was higher than the bias in original prediction. Asymptotically, the bias in original prediction converged to zero at a faster rate than the bias in switched prediction. The influence of this switched bias term was further amplified through the additional bias term $\delta_j$. Recall that the term $\delta_j$ was defined as:
	
	\begin{equation}
		\delta_j = 2\mathbb{E}_{X_j}\mathbb{E}_{\boldsymbol{X}_{-j}}\left(f_0(\boldsymbol{X}_{j}^{\prime}) - f_0(\boldsymbol{X})\right)\left(f_0({\boldsymbol{X}_j^{\prime}}) - \mathbb{E}_{\mathcal{T}}(\hat{f}(\boldsymbol{X}_{j}^{\prime}))\right)
		\label{delta}
	\end{equation}
	where $\boldsymbol{X}_{j}^{\prime}$ is a similar random vector to $\boldsymbol{X}$, but with the $j$th component switched out with its independent replicate. The term $\delta_j$ is a product of two terms: $\left(f_0(\boldsymbol{X}_{j}^{\prime}) - f_0(\boldsymbol{X})\right)$ and $\left(f_0({\boldsymbol{X}_j^{\prime}}) - \mathbb{E}_{\mathcal{T}}(\hat{f}(\boldsymbol{X}_{j}^{\prime}))\right)$. The first term represents the change in the conditional expectation function $\mathbb{E}(Y \mid \boldsymbol{X})$ when $X_j$ is switched with its independent replicate. The second term represents the bias in the prediction function after the switch. Ideally, when an unbiased prediction function is fitted, then $\mathbb{E}_{\mathcal{T}}(\hat{f}(\boldsymbol{X}_{j}^{\prime}))\to f_0({\boldsymbol{X}_j^{\prime}})$, implying $\delta_j \to 0$. If $X_j$ is not an important predictor of $Y$, i.e., $\mathbb{E}(Y \mid X_j, \boldsymbol{X}_{-j}) = \mathbb{E}(Y \mid \boldsymbol{X}_{-j})$, then $f_0(\boldsymbol{X}_{j}^{\prime}) = f_0(\boldsymbol{X})$, and the first term of $\delta_j$ is 0, resulting in $\delta_j = 0$. In the presence of high correlation between two or more variables, the $\delta_j$ contributes to the biased estimation of MVIM, as well as the differences in the bias components impact the estimation of MVIM. \\
	
	When $X_j$ was an important predictor for $Y$, the first term of $\delta_j$ (i.e., $f_0(\boldsymbol{X}_{j}^{\prime}) - f_0(\boldsymbol{X})$) was substantial  in terms of absolute value, since switching the values of $X_j$ changes the prediction. The second term $\left(f_0({\boldsymbol{X}_j^{\prime}}) - \mathbb{E}_{\mathcal{T}}(\hat{f}(\boldsymbol{X}_{j}^{\prime}))\right)$ then became a crucial part in estimating MVIM. The Oracle model defined in Section \ref{sec.Sim} was an unbiased prediction model. That implied that when the prediction function $\hat{f}$ was estimated using the Oracle model, then $f_0({\boldsymbol{X}_j^{\prime}}) = \mathbb{E}_{\mathcal{T}}(\hat{f}(\boldsymbol{X}_{j}^{\prime}))$, further implying that $\delta_j = 0$. However, black-box models such as XGBoost provided approximately unbiased predictions, or $\lim\limits_{n\to\infty}\mathbb{E}_{\mathcal{T}}(\hat{f}(\boldsymbol{X})) = f_0(\boldsymbol{X})$, where $n$ represents the training size. The predictions of small training sizes could have large biases, especially when the relationship between the predictors and the outcome was nonlinear and non-additive. For small training sizes, both $\text{Bias}^2 \text{(switch)}$ and $\text{Bias}^2 \text{(orig)}$ were substantially large, resulting in a large absolute difference between $\left(f_0({\boldsymbol{X}_j^{\prime}})\right)$ and $\mathbb{E}_{\mathcal{T}}(\hat{f}(\boldsymbol{X}_{j}^{\prime}))$. In the case of completely independent predictors, this difference decreased with increasing training size and $\delta_j$ approached zero as observed in Table \ref{tab4.1} and Figure \ref{Fig4.1}. The convergence of $\delta_j \to 0$ slowed down in the case of a simple correlation scenario, especially for $X_1$, as observed in Table \ref{tab4.2}. The high correlation between $X_1$ and $X_5$ suggested that when one of the variables was switched, the observed pair was far from the training dataset. In this case, MVIM estimation relied on the extrapolation ability of XGBoost. \\
	
	For a better understanding, consider the following simple hypothetical example: Let the outcome $Y$ represents salary, which depends on two predictors $X_1 =$ experience and $X_5 =$ age. Assume an investigator is interested in estimating MVIM for experience using a validation set after fitting XGBoost on a large training set, and XGBoost learned the true conditional expectation function fairly accurately, thus $\mathbb{E}_{\mathcal{T}}(\hat{f}(\boldsymbol{X})) \approx f_0(\boldsymbol{X})$. A subject in the validation set is 40 years old and has 20 years of experience. A second subject is 70 years old and has 45 years of experience. While switching, the first subject's experience is switched with the second subject's experience. The switched observation now has an age of 40 years and experience of 45 years. Since this combination was never observed in the training set (even if the size of the training set is substantially large), a non-parametric prediction method is most likely going to predict the first subject's salary as the average salary of 40-year-old subjects or as the average salary of individuals with 45 years of experience. The switched data point is not available in the training set and also lies in a low-density region in the predictor space. This results in $\mathbb{E}_{\mathcal{T}}(\hat{f}(\boldsymbol{X}_{j^{\prime}}))$ having a large absolute distance from $f_0(\boldsymbol{X}_{j^{\prime}})$, resulting in a large value of $\delta_j$. This phenomenon can be observed in Table \ref{tab4.2}, where $\text{Bias}^2 \text{(switch)}$ is much larger than $\text{Bias}^2 \text{(orig)}$ for both variables for all training sizes. In this case, $X_1$ had a larger $\delta$ compared to $X_5$, since $X_1$ also had interaction with $C_1$. \\
	
	This underestimation problem was amplified when $X_1$ was moderately correlated with multiple variables. In the simulation, $X_1$ was moderately correlated with five other variables. The conditional variance $\mathbb{V}(X_1 \mid C_1, X_2, X_3, X_4, X_5) = 0.07$ was set to be a very small number. In such cases of complex correlation structure, the low-density regions became more prevalent while switching, and thus $\mathbb{E}_{\mathcal{T}}(\hat{f}(\boldsymbol{X}_{j^{\prime}}))$ was equivalent to $\mathbb{E}_{\mathcal{T}}(\hat{f}(\boldsymbol{X}))$ and thus approached $f_0(\boldsymbol{X})$ with increasing training size when estimated using XGBoost or other black-box models. This problem did not exist when the true parametric model (i.e., the Oracle model) was fitted, since in that case, the training model could provide good extrapolation.
	
	\subsection{Violation of the Positivity Assumption} 
	As demonstrated in section \ref{mvim}, subsection \ref{marg.gvim.cause}, MVIM can be represented as a causal parameter under conditional ignorability and the positivity assumption. In the presence of simple to moderate correlation between multiple predictors, we may encounter a near violation of the positivity assumption. Violating the positivity assumption increases bias in the estimates of causal parameters, with or without increasing variance \citep{petersen2012diagnosing}. When the average treatment effect (ATE) is the causal estimand of interest with a continuous treatment, the estimate can be adjusted based on the conditional density of the exposure, under some strong assumptions \citep{austin2019assessing}. In this research, MVIM is a parameter with a quadratic form. To investigate this violation of the positivity assumption the focus should be on the conditional variance of the exposure variable. The near violation of the positivity assumption for an exposure $X_j$ can be easily verified using:
	
	\begin{equation}
		R^2_{X_j} = 1 - \dfrac{\mathbb{V}(X_j\mid \boldsymbol{X}_{-j})}{\mathbb{V}(X_j)}
		\label{R2X}
	\end{equation}
	A large value of $R^2_{X_j}$ indicates a near violation of the positivity assumption, while $R^2_{X_j} \approx 1$ indicates a violation of the positivity assumption. The variance of $X_j$, denoted as $\mathbb{V}(X_j)$, can be easily estimated by calculating the empirical variance from the full sample set. However, to estimate $\mathbb{V}(X_j\mid \boldsymbol{X}_{-j})$, one needs to fit a prediction model on $X_j$ utilizing rest of the predictors $\boldsymbol{X}_{-j}$. If the positivity assumption is nearly violated, MVIM estimates based on XGBoost will be biased, highlighting the importance of checking the positivity assumption before proceeding with estimating the MVIM.
	
	\section{Conditional Variable Importance Metric (CVIM)}
	
	Returning to the definition of ``importance," the question arises whether MVIM was an appropriate metric for defining the importance of a predictor when that predictor can be explained by multiple other predictors. The MVIM was defined under the assumption that switching maintained the marginal distribution of the predictor of interest as it was defined by generalizing the VIM. \citet{strobl2008conditional, debeer2020conditional} referred to VIM as marginal importance. VIM-type importance measures inherently could not assess the importance of a predictor conditional on other predictors. As we observed from the previous sections, MVIM estimates were biased when estimated from XGBoost in the presence of correlation between predictors. Previous research reported that marginal importance measures, such as VIM and MR, were biased when estimated from black-box models like Random Forests \citep{strobl2008conditional, debeer2020conditional} and neural networks \citep{hooker2021unrestricted}. It was crucial to develop a method based on MVIM to eliminate this correlation bias. \citet{strobl2008conditional} suggested that in the presence of predictor correlation, one should consider conditional permutation of the predictor of interest instead of simple permutation when calculating VIM. They termed this new definition \emph{``conditional permutation-based importance''} or \emph{``partial importance''}. Conditional permutation also limits the predictor space and, thus, the low-density regions in the prediction space, partially addressing the extrapolation problem of black-box models. This conditional importance concept can be easily extended to conditional switching in the context of the present research. In this section, we derived a new conditional Variable Importance Metric (CVIM). This definition of CVIM relied on conditional switching of the predictor of interest. \\
	
	We defined the CVIM based on the work of \citet{strobl2008conditional, hooker2021unrestricted, fisher2019all}. Recall $O^{(a)} = (Y^{(a)}, X^{(a)}, \boldsymbol{Z}^{(a)})$ and $O^{(b)} = (Y^{(b)}, X^{(b)}, \boldsymbol{Z}^{(b)})$ be two random vectors from the same population, $f_0$ is the true conditional expectation function. An additional assumption here is that $\boldsymbol{Z}^{(a)}=\boldsymbol{Z}^{(b)} =  \boldsymbol{Z}$. Thus, $O^{(a)}$ and $O^{(b)}$ are conditionally independent given $\boldsymbol{Z}^{(a)}=\boldsymbol{Z}^{(b)} =  \boldsymbol{Z}$. Then the $e_{\text{orig}}$ is defined as
	\begin{equation}
		e_{\text{orig}}= \mathbb{E}_{X, \boldsymbol{Z}, Y}\left(Y^{(a)} - f_0( X^{(a)}, \boldsymbol{Z})\right)^{2}
	\end{equation}
	The switched error is now defined based on conditional switching. \citet{fisher2019all} termed it as $e_{\text{cond}}(f_0)$
	%	\begin{equation}
		%		e_{\text{cond}}(f_0) = \mathbb{E}_{Z ,X^{(a)}, X^{(b)} ,Y^{(b)}_{X^{(b)}}} \left[ Y_{X^{(b)}}^{(b)}-\mathbb{E}\left(Y_{X^{(a)}}^{(b)}\mid Z^{(a)}=Z^{(b)} =  Z\right) \right]^2
		%		\label{cond.switch}
		%	\end{equation}
	\begin{equation*}
		e_{\text{cond}} =\mathbb{E}_{\boldsymbol{Z}}\mathbb{E}_{X^{(b)}\mid \boldsymbol{Z}}\mathbb{E}_{X^{(a)}\mid \boldsymbol{Z}}\mathbb{E}_{Y\mid X^{(a)}, \boldsymbol{Z}}\left(Y^{(a)} -  f_0( X^{(b)}, \boldsymbol{Z})\right)^{2}
		\label{cond.switch}
	\end{equation*}
	That is $X$ is only permuted when $\boldsymbol{Z}^{(a)}=\boldsymbol{Z}^{(b)} =  \boldsymbol{Z}$ and thus we define the CVIM as,
	\begin{equation}
		\mathcal{CI}_X= e_{\text{cond}}- e_{\text{orig}}
		\label{cgvim}
	\end{equation}
	
	\subsection{CVIM as Causal Parameter}
	
	One advantage of MVIM was that, under the positivity and conditional ignorability assumptions, it can be defined as a function of CATE. The correlation bias also impacted the causal interpretation of MVIM due to a near violation of the positivity assumption. It was not clear from previous research \citep{fisher2019all, hooker2021unrestricted} whether the conditional importance measures still retained the causal interpretation of MVIM. The CVIM that we have discussed here has similar causal interpretation as MVIM. The following theorems showed that CVIM could also be represented as a function of CATE.

	\begin{theorem}
		\label{stat.proof.4.1}
		Let the treatment/exposure $X$ be a binary variable ($X \in \{0,1\}$). With respect to the true conditional expectation $f_{0}$, Equation \eqref{cgvim} can be expressed as,
		
		\begin{equation}
			\small
			\begin{split}
				\mathcal{CI}_X&= e_{\text {cond}} -  e_{\text {orig}}\\
				&= 2 \mathbb{E}_{\boldsymbol{Z}}\left(\mathbb{V}(X\mid \boldsymbol{Z})\left[\mathbb{E}(Y_1\mid \boldsymbol{Z}) - \mathbb{E}(Y_0 \mid \boldsymbol{Z})\right]^2\right)
			\end{split}
			\label{cgvim.bin}
		\end{equation}
	\end{theorem}
	
	\begin{theorem}
		\label{stat.proof.4.2}
		Let's the treatment/exposure $X$ be multinomial with $K$ categories ($x \in \{1, 2, ..., K\}$). Now with respect to the true conditional expectation $f_{0}$  Equation \eqref{cgvim} can be re-written as,
		
		\begin{equation}
			\mathcal{CI}_X =2\mathbb{E}_{\boldsymbol{Z}}\left(\sum_{x = 1}^{K}\sum_{x^* \neq x}p_{x}(\boldsymbol{Z})p_{x^*}(\boldsymbol{Z})\left[\mathbb{E}(Y_x \mid \boldsymbol{Z}) - \mathbb{E}(Y_{x^*}\mid\boldsymbol{Z})\right]^2 \right) 
			\label{cgvim.mult}
		\end{equation}
		where, $p_{x}(\boldsymbol{Z}) = P(X = x\mid \boldsymbol{Z})$ and $p_{x^*}(\boldsymbol{Z}) = P(X = x^*\mid \boldsymbol{Z})$. Again, $Y_x$ and $Y_{x^*}$ represent the potential outcomes for treatment level $X = x$ and $X = x^*$ respectively.
		
	\end{theorem}

	\begin{theorem}
		\label{stat.proof.4.3}
		For a continuous treatment/exposure $X$, the $\mathcal{CI}$ can be expressed as,
		\begin{equation}
				\mathbb{E}_{\boldsymbol{Z}}\left[\int_{X}\left[\int_{X^*} \left[\mathbb{E}\left(Y_{x}\mid \boldsymbol{Z}\right) - \mathbb{E}\left(Y_{x^*} \mid  \boldsymbol{Z}\right)\right]^2 dP_{X^*\mid \boldsymbol{Z}}(x^*)\right] \right.
				\left. dP_{X\mid \boldsymbol{Z}}(x)\right]
			\label{cgvim.cont}
		\end{equation}
		Here, $x$ and $x^*$ represent the treatment levels under two counterfactual scenarios.
	\end{theorem}

	\subsection{CVIM and Causal Variance Decomposition}
	The MVIM and the CVIM were defined based on mean squared errors, which represents how much the conditional variance of the outcome $Y$ changes after switching the predictor or a the treatment of interest. In a causal context, this can be interpreted as how much MVIM or CVIM can causally explain the total variation of the outcome for a treatment of interest, which can be obtained via a variance decomposition. This variance decomposition was referred to as the causal variance decomposition by \citet{chen2020causal} who illustrated the causal variance decomposition for binary and multinomial exposures. They showed that the marginal variance of the response $Y$ can be decomposed as,
	
	\begin{equation}
		\begin{split}
			\mathbb{V}(Y) & = \mathbb{V}_{Z}\left(\sum_{k\in \{0,1\}}\mathbb{E}(Y_k\mid Z)p_k(Z)\right)\\
			& + \mathbb{E}_{Z}\left(\mathbb{V}(X\mid Z)\left[\mathbb{E}(Y_1\mid Z) - \mathbb{E}(Y_0 \mid Z)\right]^2\right)\\
			&	+ \mathbb{E}_{Z}\left(\sum_{k\in \{0,1\}}\mathbb{V}(Y_k)p_k(Z)\right)
		\end{split}
		\label{causal.var}
	\end{equation}
	where $X$ is a binary exposure and $Y_k$ is the potential outcome corresponding to the exposure level $X = k$. \citet{chen2020causal} defined the second term in the decomposition as \emph{average variance causally explained by the between exposure level differences conditional on the confounders}. They further derived the second term as, $\mathbb{E}_{X,Z}\left\{\left[\mathbb{E}(Y\mid X, Z) - \mathbb{E}(Y\mid Z)\right]^2\right\}$, which is the same as the LOCO estimator defined in the equation \eqref{loco}. That is, the second term is the half of the CVIM for a binary exposure. If the positivity assumption is violated, then the second term equates to zero, implying that there is no causal variation. CVIM can directly estimate the second term, and lower values of CVIM may indicate either that the variable is not important or that there is a near violation of the positivity assumption. Similar decomposition can also be expressed for a multinomial exposure. To our knowledge, this is the first comparison of the conditional permutation-based importance and the causal variance decomposition. The advantages CVIM has over previously defined methods \citep{chen2020causal} are that CVIM can be estimated for any type of treatment, including continuous treatments, and also the estimation technique does not rely on any parametric assumptions.
	
	\subsection{Estimation of CVIM}
	\label{CGVIM.mult.imp}
	Given that the confounders vector is defined as $\boldsymbol{Z} = \boldsymbol{X}_{-j}$, where $\boldsymbol{X}_{-j}$ may consist of a large number of predictors, performing a conditional permutation of $X = X_j$ given $\boldsymbol{Z} = \boldsymbol{X}_{-j}$ becomes non-trivial. To address this challenge, we propose estimating CVIM by replacing the values of a predictor $X_j$ in the validation set with observations simulated from the conditional density $g(X_j\mid \boldsymbol{X}_{-j})$. Specifically, the process involves first fitting a prediction model of $X_j$ on $\boldsymbol{X}_{-j}$ to determine the conditional density $g(X_j\mid \boldsymbol{X}_{-j})$, and then using this density to perform the replacement. In a simplified scenario where one can assume constant variance for a continuous predictor $X_j$ given $\boldsymbol{X}_{-j}$, one can define:
	\begin{equation}
		X_j = \mathbb{E}(X_j\mid \boldsymbol{X}_{-j}) + \nu
	\end{equation}
	Here, $\nu$ has a mean of 0 and variance $\sigma^2_\nu$. Predicting the conditional expectation $\mathbb{E}(X_j\mid \boldsymbol{X}_{-j}) $ can be achieved using any machine learning model, while $\sigma^2_\nu$ can be estimated by computing prediction errors from a validation set. Once these estimates are obtained, the values of $X_j$ in the validation set can be switched by randomly generating observations from the conditional density of $X_j \mid \boldsymbol{X}_{-j}$, while estimating $e_{\text{cond}}(f_0)$. This can be executed by computing residual as $r_j = X_j -  \hat{X}_{j}$, where $\hat{X}_j$ is predicted from any machine learning model of interest. To estimate $e_{\text{cond}}$, the prediction error is recalculated by permuting only the residuals and adding them to $\hat{X}_{j}$. Once the switched values are obtained for $X_j$, the $e_{\text{orig}}$ can be estimated as,
	\begin{equation}
		\hat{e}_{\text{orig}} = \dfrac{1}{n_{v}}\sum_{i =1}^{n_v}\left(Y - \hat{f}(X_j, \boldsymbol{X}_{-j})\right)^2
	\end{equation}
	where, $n_v$ is the number of observations in the validation set and $\hat{f}(.)$ is the prediction function. The $e_{\text{cond}}$ can be estimated with,
	\begin{equation}
		\hat{e}_{\text{cond}} = \dfrac{1}{n_{v}}\sum_{i =1}^{n_v}\left(Y - \hat{f}(\hat{X}_j + r_j^{\prime}, \boldsymbol{X}_{-j})\right)^2
	\end{equation}
	where, $r_j^{\prime}$ represents the switched residuals. Finally, CVIM is estimated with $\widehat{\mathcal{CI}} = \hat{e}_{\text{cond}}  - \hat{e}_{\text{orig}}$.\\
	
	Again, if $X_j$ is binary, $\mathbb{E}(X_j\mid \boldsymbol{X}_{-j})  = P(X_j\mid \boldsymbol{X}_{-j} = \boldsymbol{x}_{-j})$ can be estimated, and the switched values of $X_j$ can be generated from Bernoulli trials with this conditional probability. Similarly, if $X_j$ is multinomial, the conditional probability of the $X = x$ can be estimated, and the switched value of $X_j$ can be generated from a multinomial distribution using the probabilities. The algorithm of estimating CVIM is as follows.
	
	\begin{algorithm}[H]
		\caption{CVIM Estimation}\label{alg.cgvim}
		\begin{algorithmic}[1]
			\FOR{B = $1$ to $100$}
			\STATE Obtain a Bootstrap sample from the full dataset
			\FOR{k = $1$ to $10$}
			\STATE Split the dataset into training (67\%) and validation (33\%) sets
			\STATE Fit a ML model to the response variable using the all the predictors training dataset.
			\STATE Predict $\hat{f}$ the change scores $y$ from the validation set 
			\STATE calculate $\hat{e}_{\text{orig}} = \frac{1}{n_{\text{vald}}}\sum_{i=1}^{n_{\text{vald}}}\left(y_i - \hat{f}_i\right)^2$
			\STATE Fit another ML model to the risk factor using the training dataset which all the confounders but not the response (change).
			\FOR{m in $1$ to $5$}
			\STATE Conditionally Permute the predictor of interest in the validation set. 
			\STATE Recalculate the predictions after permuting as $\hat{f}_{j^{\prime}}$
			\STATE calculate $\hat{e}_{\text{cond}} = \frac{1}{n_{\text{vald}}}\sum_{i=1}^{n_{\text{vald}}}\left(y_i - \hat{f}_{j^{\prime}, i}\right)^2$
			\STATE Calculate $\mathcal{CI}_{mkB} = \hat{e}_{\text{cond} } - \hat{e}_{\text{orig}}$
			\ENDFOR
			\ENDFOR
			\ENDFOR
			\STATE Estimate $\widehat{\mathcal{CI}}$ by averaging the $\mathcal{CI}_{mkB}$s
		\end{algorithmic}
	\end{algorithm} 
	
	\subsection{Bias-Variance Decomposition of CVIM}
	
	In the next step we derived the bias-variance decomposition for CVIM. The decomposition for $\hat{e}_{\text{orig}}$ is the same as the one derived in equation  \eqref{BV1}. For the decomposition of $\hat{e}_{\text{cond}}$ the only difference was the order of expectation as the CVIM is derived based on the conditional expectation $\mathbb{E}(X_j \mid \boldsymbol{X}_{-j})$. The bias-variance terms for $e_{\text{cond}}$ are as follows,
	%\vspace{-10pt}
	\begin{equation}
		\small
		\begin{split}
			%		\begin{align}
				& \mathbb{E}_{ \boldsymbol{X}_{-j}}\mathbb{E}_{X_j\mid \boldsymbol{X}_{-j}}\mathbb{E}_{Y\mid \boldsymbol{X}}\left(\hat{e}_{\text{cond}}\right) 	\\
				& = \mathbb{E}_{\boldsymbol{X},Y}\left(Y - \hat{f}(\boldsymbol{X}^{\prime}_j)\right)^2 \notag \\
				%			& = \mathbb{E}_{\boldsymbol{X},Y}\left(Y - f_0(\boldsymbol{X}^{\prime}_j)\right)^2 +  	\mathbb{E}_{\boldsymbol{X}}\left(\mathbb{E}_{\mathcal{T}}(\hat{f}(\boldsymbol{X}^{\prime}_j)) -  f_0(\boldsymbol{X}^{\prime}_j)\right)^2 \notag \\
				%			& + \mathbb{E}_{\boldsymbol{X}}\mathbb{E}_{\mathcal{T}}\left(\hat{f}(\boldsymbol{X}^{\prime}_j) - \mathbb{E}_{\mathcal{T}}(\hat{f}(\boldsymbol{X})) \right)^2 \notag \\
				%			& + 2\mathbb{E}_{ \boldsymbol{X}_{-j}}\mathbb{E}_{X_j\mid \boldsymbol{X}_{-j}}\left(f_0(\boldsymbol{X} )- f_0(\boldsymbol{X}_{j}^{\prime}))\right)\left(f_0({\boldsymbol{X}_j^{\prime}}) - \mathbb{E}_{\mathcal{T}}(\hat{f}(\boldsymbol{X}_{j}^{\prime})) \right)  \notag \\ 
				& = e_{\text{cond}} +  	\underbrace{\mathbb{E}_{ \boldsymbol{X}_{-j}}\mathbb{E}_{X_j\mid \boldsymbol{X}_{-j}}\left(\mathbb{E}_{\mathcal{T}}(\hat{f}(\boldsymbol{X}_{j}^{\prime})) -  f_0(\boldsymbol{X}_{j}^{\prime})\right)^2}_{\text{Bias}^2\text{(cond)}} \notag \\
				& + \underbrace{\mathbb{E}_{ \boldsymbol{X}_{-j}}\mathbb{E}_{X_j\mid \boldsymbol{X}_{-j}}\mathbb{E}_{\mathcal{T}}\left(\hat{f}({\boldsymbol{X}_j^{\prime}}) - \mathbb{E}_{\mathcal{T}}(\hat{f}(\boldsymbol{X}_{j}^{\prime})) \right)^2}_{\text{Variance (cond)}} \notag \\
				& +\underbrace{2\mathbb{E}_{ \boldsymbol{X}_{-j}}\mathbb{E}_{X_j\mid \boldsymbol{X}_{-j}}\left(f_0(\boldsymbol{X}_{j}^{\prime}) - f_0(\boldsymbol{X} )\right)\left(f_0({\boldsymbol{X}_j^{\prime}}) - \mathbb{E}_{\mathcal{T}}(\hat{f}(\boldsymbol{X}_{j}^{\prime}) \right)}_{\text{additional bias term }\delta_j}
				%	\end{align}
			\label{BV.cond}
		\end{split}
	\end{equation}
	here, $\boldsymbol{X}_{j}^{\prime}$ is the vector where the $j$th column is conditionally switched as defined in equation\eqref{cond.switch}. The estimated CVIM can be decomposed as
	
	\begin{equation}
		\mathbb{E}(\widehat{\mathcal{CI}}_j)  \approx \mathcal{CI}_j + \delta_j
		\label{BV.cond2}
	\end{equation}
	where, $ \mathcal{CI}_j$ is the true CVIM for the predictor $X_j$. Like the bias-variance decomposition of MVIM, the bias-variance decomposition of CVIM also includes a $\delta_j$ term. The definition of $\delta_j$ is slightly altered based on the order of expectation over $\boldsymbol{X}$. A high absolute value of the additional term $\delta_j$ can induce bias in CVIM estimation even if the assumptions regarding the equity of the bias and variance terms are true. To investigate further we designed another simulation study to evaluate the behavior of the term $\delta_j$ and the overall bias in estimating CVIM in several different scenarios. 
	
	\subsection{Simulation Results of CVIM}
	To illustrate the statistical properties of CVIM, we conducted simulations based on the procedure described in subsection \ref{sec.Sim}. The outcome $Y$ is generated using equation \eqref{sim.mars} and $X_1$ was generated using equation \eqref{x.gen} as described in the multivariate correlation scenario in the subsection \ref{mult.cor}. In this section we present three scenarios based on the dependence of $X_1$ on rest of the predictors.
	\noindent	\textbf{Weak Dependence} In this scenario $X_1$ was generated using
	\begin{equation}
		\begin{split}
			X_1 & =  -0.5 + C_1 -0.5X_2 + 0.5X_3 + 0.3X_4 - 0.3X_5  + \nu  \\
			& \text{ }\nu \sim N(0,\sigma^2_\nu = 1^2 = 1)
		\end{split}
		\label{x.gen.v1}
	\end{equation}
	here, $\mathbb{V}(X_1) = \mathbb{V}(C_1 -0.5X_2 + 0.5X_3 + X_4 - X_5  + \nu) = 1.93$ and $\mathbb{V}(X_1 \mid \boldsymbol{X}_{-1}) = 1$. Thus, as defined in equation \eqref{R2X}, the $R^2_{X_1} = 0.48$.  This scenario is defined as the \emph{weak dependence:} of $X_1$ on the other predictors. 
	%\newpage
	\noindent\textbf{Moderate dependence:} In this scenario $X_1$ was generated using
	\begin{equation}
		\begin{split}
			X_1  & =  -0.5 + C_1 -0.5X_2 + 0.5X_3 + 0.3X_4 - 0.3X_5  + \nu  \\
			& \text{ }\nu \sim N(0,\sigma^2_\nu = 0.5075^2 = 0.2576)
		\end{split}
		\label{x.gen.v2}
	\end{equation}
	here, $\mathbb{V}(X_1) = \mathbb{V}(C_1 -0.5X_2 + 0.5X_3 + X_4 - X_5  + \nu) = 1.188$ and $\mathbb{V}(X_1 \mid \boldsymbol{X}_{-1}) = 0.2576$. The $R^2_{X_1} = 0.78$.  This scenario is defined as the \emph{moderate dependence} of $X_1$ on the other predictors. \\
	\textbf{Strong dependence:} In this scenario $X_1$ was generated using
	\begin{equation}
		\begin{split}
			X_1  &=  -0.5 + C_1 -0.5X_2 + 0.5X_3 + 0.3X_4 - 0.3X_5   + \nu\\
			& \text{ }\nu \sim N(0,\sigma^2_\nu = 0.2575^2 = 0.066)
		\end{split}
		\label{x.gen.v3}
	\end{equation}
	here, $\mathbb{V}(X_1) = \mathbb{V}(C_1 -0.5X_2 + 0.5X_3 + X_4 - X_5  + \nu) = 1$ and $\mathbb{V}(X_1 \mid \boldsymbol{X}_{-1}) = 0.066$. The $R^2_{X_1} = 0.93$.  This scenario is defined as the \emph{strong dependence} of $X_1$ on the other predictors. This scenario can be considered as a near violation of the positivity assumption.

	\subsection{Plots of Bias Variance Components}
	
	We generated training sets of various sizes to train the prediction model. We further generated a large validation set of size of 100,000 to calculate the true CVIMs for $X_1$ using Monte Carlo method. Using 100 simulations the bias-variance components were calculated for each observations of the large validation set and then the average of the bias-variance components were then calculated utilizing all the observations of the large validation set. The simulations were conducted for three scenarios as described in the previous section.
	
	\subsubsection{Weak dependence}
	
	The bias variance components for $X_1$ are as follows
	\begin{table}[ht]
		\centering
		\caption{Components of bias-variance decomposition of $\widehat{\mathcal{CI}}$, as described in \eqref{BV.cond2}. The results were obtained for predictors $X_1$}
		\begin{tabular}{rrrrrrrrr}  \hline
			& \multicolumn{2}{c}{$\text{Bias}^2$} & \multicolumn{2}{c}{$\text{Variance}$} & & \multicolumn{3}{c}{$\text{CVIM}$} \\ \hline
			$n_{\text{train}}$  & Cond & Orig & Cond & Orig  & $\delta_j$ & $\widehat{\mathcal{CI}}$ & $\mathcal{CI}_c$ & $\mathcal{CI}$ \\   \hline   \hline
			100 & 22.70 & 22.87 & 14.93 & 14.93 & 7.43 & 0.39 & 8.00 & 8.00 \\   
			500 & 8.85 & 8.90 & 10.19 & 10.19 & 5.56 & 2.39 & 8.00 & \\   
			1000 & 5.00 & 5.01 & 8.29 & 8.28 & 4.28 & 3.71 & 8.00 &  \\  
			5000 & 1.07 & 1.05 & 4.42 & 4.37 & 1.64 & 6.43 & 8.00 &  \\   
			10000 & 0.55 & 0.53 & 3.17 & 3.07 & 0.97 & 7.15 & 8.00 &  \\   
			20000 & 0.31 & 0.29 & 2.13 & 1.99 & 0.52 & 7.65 & 8.01 &  \\   
			50000 & 0.27 & 0.27 & 1.12 & 0.93 & 0.24 & 7.99 & 8.03 &  \\    
			\hline
		\end{tabular}
		\label{tab.cgvim.1}
	\end{table}
	
	\begin{figure}[H]
		\centering
		\includegraphics[scale = 0.35]{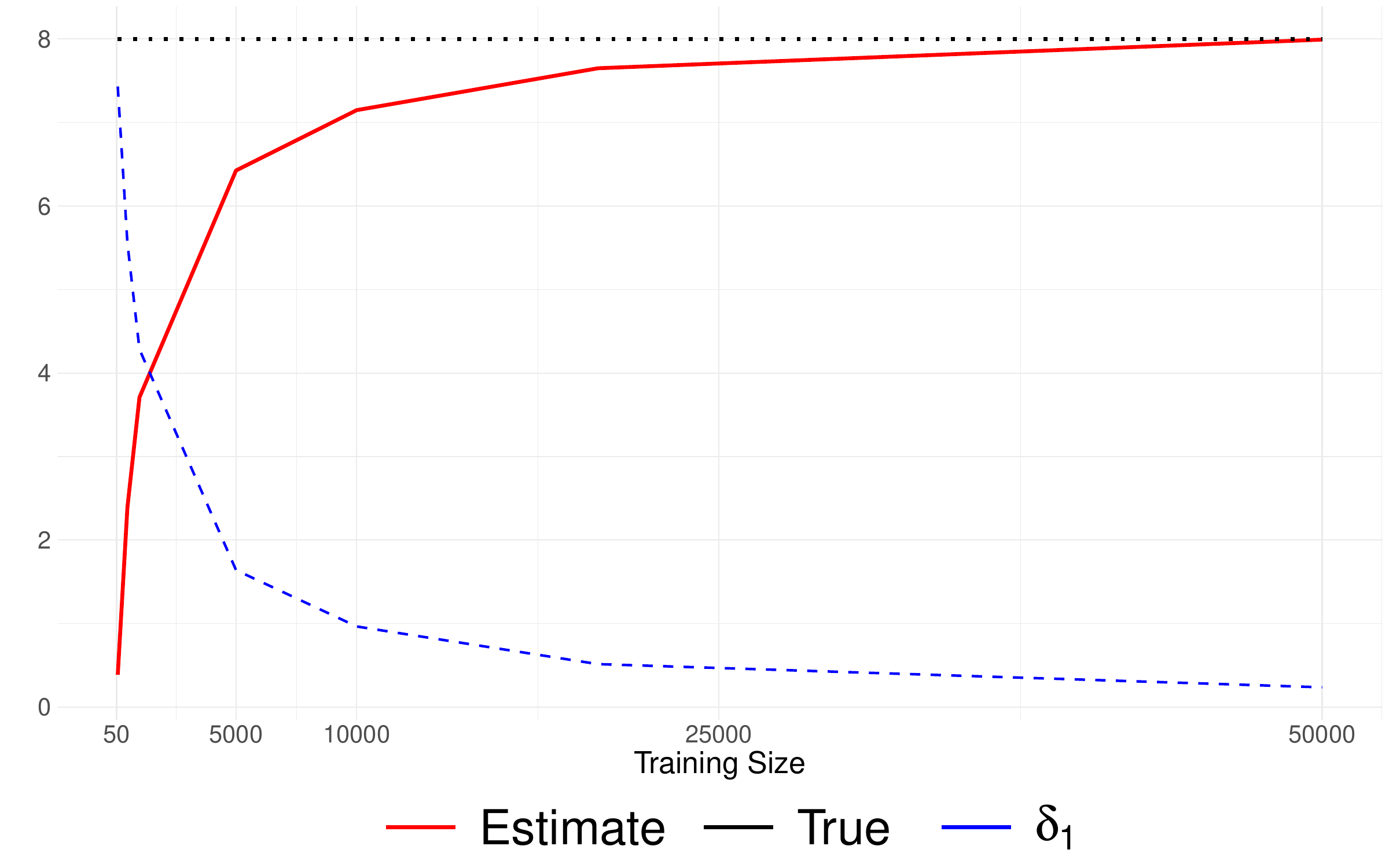}
		\caption{The trajectory of $\delta_1$ and $\widehat{\mathcal{CI}}$ of $X_1$ with increasing training size obtained from the weak dependence scenario}
		\label{Fig4.4}
	\end{figure}
	
	In this scenario the bias and variance components before and after switching were similar. The $\delta_1$ was large for smaller training sizes and thus, produced bias estimates of CVIM. With increasing training sizes the $\delta_1$ converged to zero. The estimate $\widehat{\mathcal{CI}}$ was an approximately unbiased and consistent estimate of the true CVIM.
	
	\subsubsection*{Moderate dependence}
	
	The bias variance components for $X_1$ are as follows
	\begin{table}[ht]
		\centering
		\caption{Components of bias-variance decomposition of $\widehat{\mathcal{CI}}$, as described in \eqref{BV.cond2}. The results are obtained for predictors $X_1$}
		\begin{tabular}{rrrrrrrrr}  \hline
			& \multicolumn{2}{c}{$\text{Bias}^2$} & \multicolumn{2}{c}{$\text{Variance}$} & & \multicolumn{3}{c}{$\text{CVIM}$} \\ \hline
			$n_{\text{train}}$ & Cond & Orig & Cond & Orig  & $\delta_j$ & $\widehat{\mathcal{CI}}$ & $\mathcal{CI}_c$ & $\mathcal{CI}$ \\  \hline
			100 & 19.98 & 20.05 & 13.34 & 13.35 & 1.91 & 0.07 & 2.06 & 2.06 \\  
			500 & 7.33 & 7.36 & 8.81 & 8.81 & 1.64 & 0.39 & 2.06 &  \\  
			1000 & 4.26 & 4.27 & 7.10 & 7.10 & 1.38 & 0.68 & 2.06 &  \\   
			5000 & 1.03 & 1.01 & 3.85 & 3.83 & 0.84 & 1.26 & 2.06 &  \\   
			10000 & 0.56 & 0.54 & 2.74 & 2.70 & 0.66 & 1.46 & 2.06 &  \\  
			20000 & 0.30 & 0.29 & 1.85 & 1.79 & 0.40 & 1.75 & 2.07 &  \\   
			50000 & 0.27 & 0.26 & 0.95 & 0.85 & 0.22 & 1.97 & 2.09 &  \\    
			\hline
		\end{tabular}
		\label{tab.cgvim.2}
	\end{table}
	
	\begin{figure}[H]
		\centering
		\includegraphics[scale = 0.35]{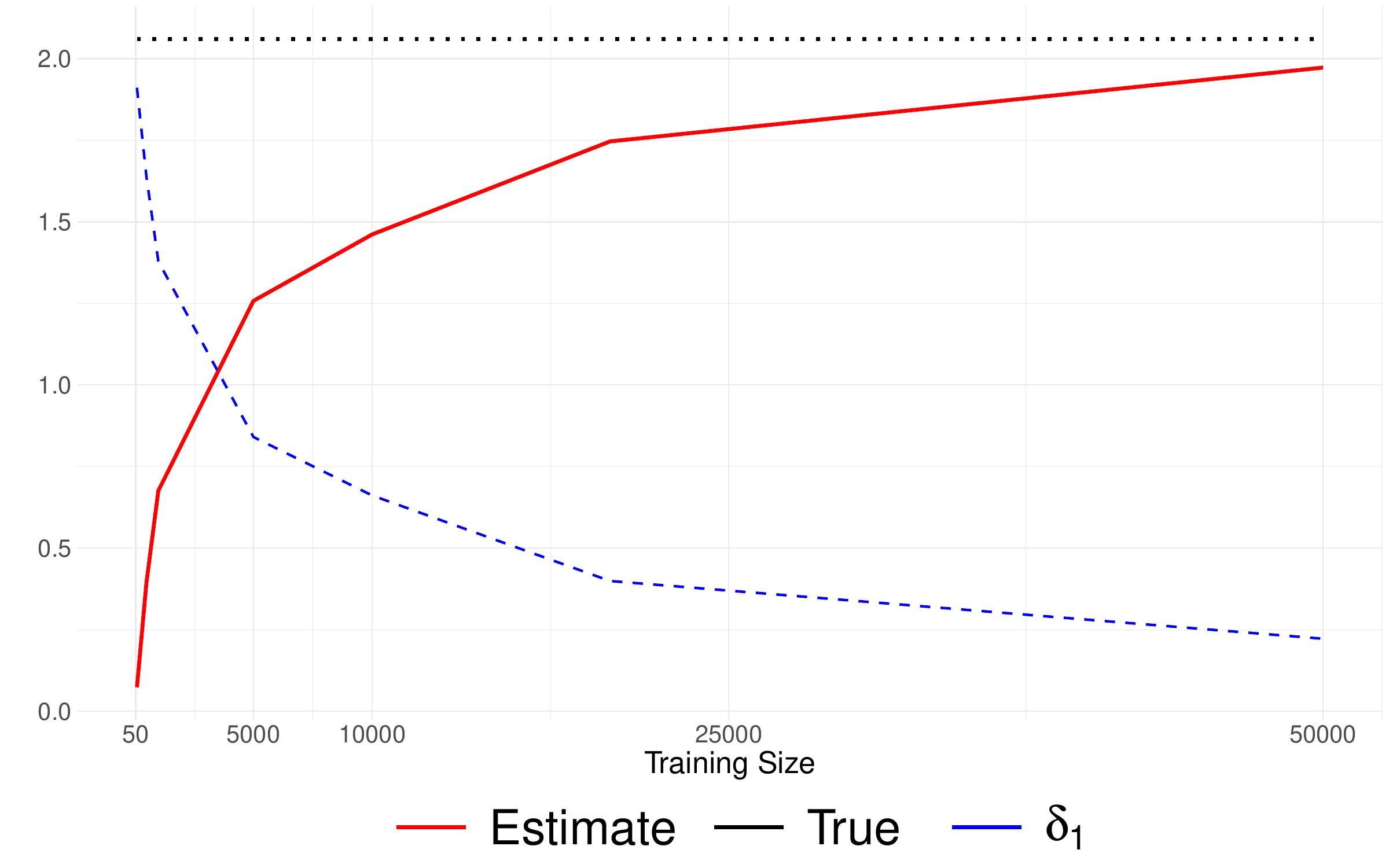}
		\caption{The trajectory of $\delta_1$ and $\widehat{\mathcal{CI}}$ of $X_1$ with increasing training size obtained from the moderate dependence scenario}
		\label{Fig4.5}
	\end{figure}
	
	In this scenario the the differences between the bias and variance components before and after switching were negligible. The values of the bias and variance components were large for smaller training set sizes but diminished with increasing training set size. Again, the $\delta_1$ was large for smaller training sizes and thus, produced biased estimates of CVIM. With increasing training sizes the $\delta_1$ converged to zero. The convergence rate was slower compared to the weak dependence scenario. The estimate $\widehat{\mathcal{CI}}$ again was an approximately unbiased and consistent estimate of the true CVIM.

	\subsubsection{Strong dependence}
	
	The bias variance components for $X_1$ are as follows
	\begin{table}[ht]
		\centering
		\caption{Components of bias-variance decomposition of $\widehat{\mathcal{CI}}$, as described in \eqref{BV.cond2}. The results are obtained for predictors $X_1$}
		\begin{tabular}{rrrrrrrrr}  \hline
			& \multicolumn{2}{c}{$\text{Bias}^2$} & \multicolumn{2}{c}{$\text{Variance}$} & & \multicolumn{3}{c}{$\text{CVIM}$} \\ \hline
			$n_{\text{train}}$  & Cond & Orig & Cond & Orig  & $\delta_j$ & $\widehat{\mathcal{CI}}$ & $\mathcal{CI}_c$ & $\mathcal{CI}$ \\  \hline
			100 & 19.07 & 19.10 & 13.09 & 13.10 & 0.48 & 0.01 & 0.53 & 0.53 \\   
			500 & 6.82 & 6.83 & 8.34 & 8.34 & 0.44 & 0.08 & 0.53 &  \\   
			1000 & 3.92 & 3.92 & 6.73 & 6.73 & 0.41 & 0.12 & 0.53 &  \\   
			5000 & 0.94 & 0.92 & 3.63 & 3.62 & 0.34 & 0.21 & 0.53 &  \\   
			10000 & 0.54 & 0.52 & 2.61 & 2.60 & 0.31 & 0.26 & 0.53 &  \\   
			20000 & 0.32 & 0.29 & 1.75 & 1.72 & 0.27 & 0.32 & 0.54 &  \\   
			50000 & 0.28 & 0.26 & 0.89 & 0.84 & 0.20 & 0.42 & 0.55 &  \\ 
			\hline
		\end{tabular}
		\label{tab.cgvim.3}
	\end{table}
	
	\begin{figure}[H]
		\centering
		\includegraphics[scale = 0.35]{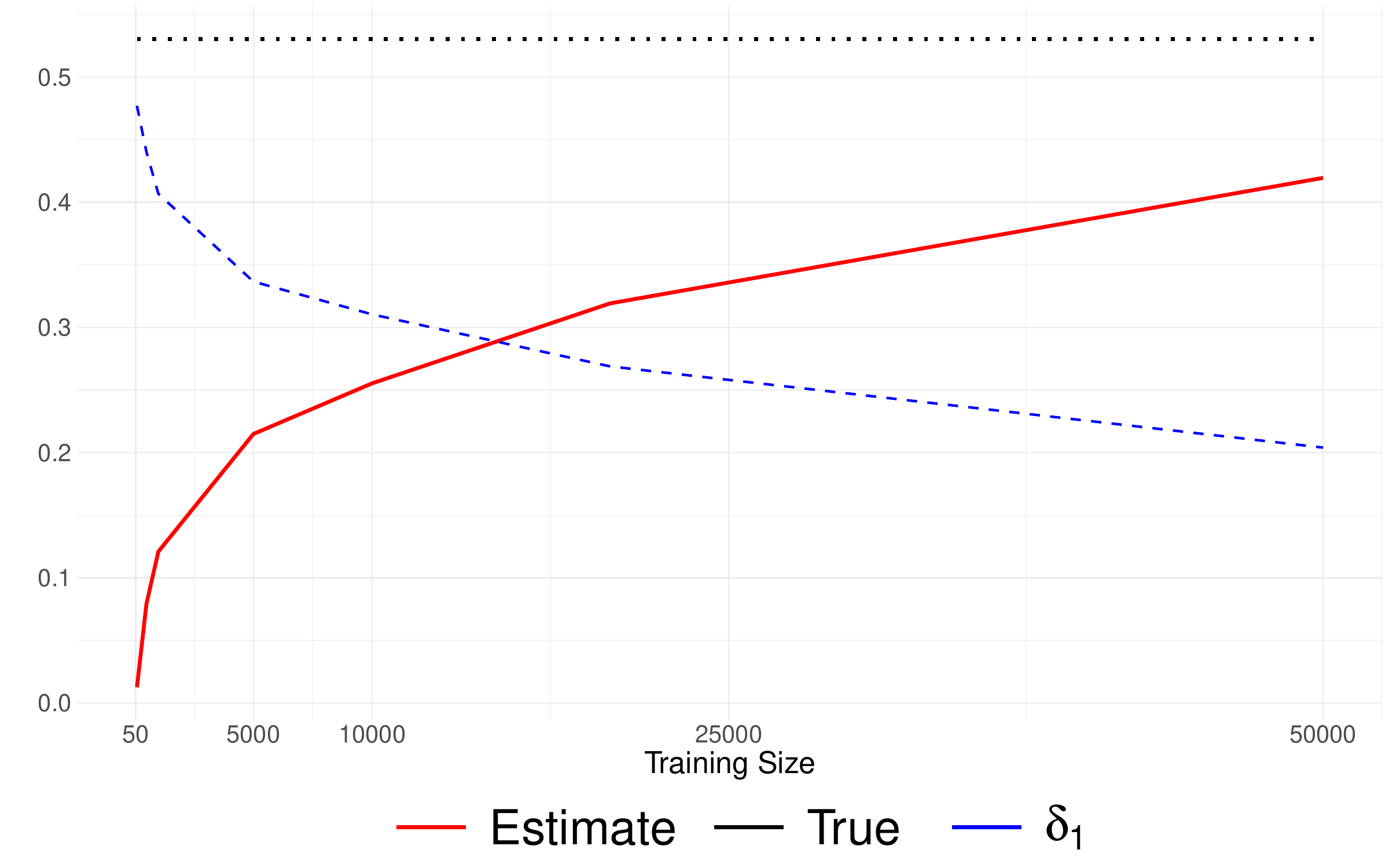}
		\caption{The trajectory of $\delta_1$ and $\widehat{\mathcal{CI}}$ of $X_1$ with increasing training size obtained from the strong dependence scenario}
		\label{Fig4.6}
	\end{figure}
	
	Again in this scenario the bias and variance components were very similar before and after switching. The $\delta_1$ was large for smaller training sizes and thus, produced biased estimates of CVIM. Here, the $\delta_1$ also converged to zero but with a definite slower rate compared to the other two scenarios. Similarly the $\widehat{\mathcal{CI}}$ also converged to the true value of CVIM but with a slower rate. Since, the correlation between the predictors are much stronger here compared to the other two scenarios. This, can be considered a scenario with a near violation of the positivity assumption and thus, the value of CVIM (0.53) is much lower to the value of MVIM (8). With stronger dependence structure (i.e., $R^2_{X_1} \rightarrow 1$), the true conditional importance would drop down to zero and thus the CVIM estimate will also be zero, however the true MVIM could be much larger, indicating the violation of the positivity assumption.
	
	\section{Relationship between MVIM and CVIM}
	
	As correlation among predictors increases, the true value of CVIM decreases, unlike the true MVIM, which is not affected by the correlation, whereas the estimated MVIM can be highly biased in presence of predictor correlation. The CVIM estimates are less affected by the correlation when compared to the MVIM estimates. Additionally, CVIM can signal a potential violation of the positivity assumption. It is evident that CVIM should be the preferred measure of importance when predictors are highly correlated, as discussed by \citet{strobl2008conditional}, \citet{fisher2019all} \citet{debeer2020conditional}, \citet{hooker2021unrestricted}, and \citep{chamma2024statistically}, given that MVIM estimates are biased in the presence of predictor correlation. This is especially relevant when $\boldsymbol{X}_{-j}$ are true confounders that causally affect both $X_j$ and $Y$. However, when $\boldsymbol{X}_{-j}$ do not affect $Y$ but are highly correlated with $X_j$, CVIM might still be very low even if $X_j$ is the only important predictor. As noted by \citet{verdinelli2024feature}, if the true conditional expectation is a linear function of the predictors and a predictor is perfectly correlated with another feature, it can be dropped without loss of prediction accuracy, resulting in LOCO/CVIM being 0, which is technically correct. However, this might lead to the misinterpretation that the model does not depend on this predictor, despite its regression coefficient being large. While parametric models can accurately estimate both MVIM and CVIM when model assumptions hold true, in practice, these assumptions are most likely to be violated. Our results indicate that black-box models perform poorly in estimating MVIM due to their inability to extrapolate in low-density regions, but they can estimate CVIM more accurately. To address this issue we investigated further on the relationship between GVIM and CVIM. If a predictor has a linear or additive term in $f_0(\boldsymbol{X})$, an analytical relationship between GVIM and CVIM can be established, as indicated in equation \eqref{vim.ad.mod}. The CVIM for an additive model can be re-written as,
	
	\begin{equation}
		\begin{split}
			\mathcal{CI}_j& = \mathbb{E}\left[  (f_{j}(X_{j}) - f_{j}(X^{\prime}_{j}))^{2} \right]  \\
			%		& = \mathbb{E}_{\boldsymbol{X}_{-j}}\mathbb{E}_{X_j \mid \boldsymbol{X}_{-j}}\left[  (f_{j}(X_{j}) - \mathbb{E}(f_{j}(X_{j})) + \mathbb{E}(f_{j}(X_{j})) + f_{j}(X^{\prime}_{j}))^{2} \right]\\
			& = 2\mathbb{E}_{\boldsymbol{X}_{-j}}\mathbb{V}_{X_j \mid \boldsymbol{X}_{-j}}\left[f_{j}(X_{j})\right]
		\end{split}
		\label{cgvim.ad.mod}
	\end{equation}
	A simple scenario is to assume homogeneity in conditional variance of $X_j$ that is, $\mathbb{V}_{X_j \mid \boldsymbol{X}_{-j}}(X_{j}) = \sigma^2_\nu$. Then we can write,
	\begin{equation}
		\mathbb{E}_{\boldsymbol{X}_{-j}}\mathbb{V}_{X_j \mid \boldsymbol{X}_{-j}}\left[f_{j}(X_{j})\right] = \mathbb{V}_{X_j \mid \boldsymbol{X}_{-j}}\left[f_{j}(X_{j})\right]
	\end{equation}
	thus, when $X_j$ has a linear term in $f_0(\boldsymbol{X})$, i.e., $f_j(X_j) = \beta_jX_j$. Then \eqref{cgvim.ad.mod},  
	
	\begin{equation}
		\begin{split}
			&	\mathcal{CI}_j = 2\beta_j^2\mathbb{V}(X_j\mid \boldsymbol{X}_{-j}) = \mathcal{MI}_j \dfrac{\mathbb{V}(X_j\mid \boldsymbol{X}_{-j}) }{\mathbb{V}(X_j)}  \\
			\implies	& \mathcal{CI}_j = \mathcal{MI}_j \dfrac{\sigma^2_\nu}{\mathbb{V}(X_j)} = \mathcal{MI}_j(1 - R^2_{X_j})\\
			\implies & \mathcal{MI}_j = \dfrac{\mathcal{CI}_j }{1- R^2_{X_j} }
		\end{split}
		\label{cgvim.lin.mod}
	\end{equation}
	This results are obtainable since $\mathcal{MI} = 2\mathbb{V}(f_j(X_j)) = 2\mathbb{V}(\beta_j X_j) = 2\beta_j^2\mathbb{V}(X_j)$. In this case the MVIM is indeterminable using CVIM when $R^2_{X_j} = 1$. The assumption of constant variance for \(X_j\) may not always be realistic. In this situation, the positivity assumption is violated only when \(\mathbb{V}(X_j \mid \boldsymbol{X}_{-j}) = 0\). This represents a specific instance of positivity violation. It is important to note that even if the conditional variance is not constant, the positivity assumption can still be violated if there exists some \(\boldsymbol{X}_{-j} = \boldsymbol{x}_{-j}\) for which \(\mathbb{V}(X_j \mid \boldsymbol{X}_{-j} = \boldsymbol{x}_{-j}) = 0\). The current set of simulation scenarios in this manuscript considers only the simple case where the variance of \(X_j \mid \boldsymbol{X}_{-j}\) is constant.
	\\ 
	
	Since MVIM estimates are unreliable when calculated from machine learning models, an \emph{adjusted MVIM (AMVIM)} can be derived using equation \eqref{cgvim.lin.mod} if the predictor of interest is assumed to have a linear relationship with the outcome. For demonstration, we simulated a dataset using equations \eqref{sim.mars}. The predictor $X_j$ was generated from three scenarios defined in equations \eqref{x.gen.v1}, \eqref{x.gen.v2}, and \eqref{x.gen.v3}. We simulated datasets of various sizes. Following the process in subsection \ref{sec.Sim}, two-thirds of the dataset was used as a training set and the remaining one-third as a validation set. First, we fitted an XGBoost model to predict $\mathbb{E}(X_j\mid \boldsymbol{X}_{-j})$ and estimated $\mathbb{V}(X_j \mid \boldsymbol{X}_{-j})$ by calculating $\hat{\sigma}^2_\nu$ from the validation sets. Next, another XGBoost model $\hat{f}(\boldsymbol{X})$ was fitted to predict the response $Y$. We estimated $e_{\text{orig}}$ by calculating the prediction error for $Y$ from the validation set. We further estimated $e_{\text{cond}}$ by first obtaining the residual vector $r_j = x_j - \hat{\mu}_{X_j}(\boldsymbol{x}_{-j})$, permuting the residuals, and recalculating the prediction error for $Y$. The marginal variance of $X_j$ was estimated by calculating the sample variance of $X_j$ from the full dataset. Finally, we estimated MVIM, CVIM and AMVIM
	\begin{enumerate}
		\item The original MVIM was estimated by $\widehat{\mathcal{MI}}_j = \hat{e}_{\text{switch}} - \hat{e}_{\text{orig}}$.
		\item The CVIM was estimated by $\widehat{\mathcal{CI}}_j = \hat{e}_{\text{cond}} - \hat{e}_{\text{orig}}$.
		\item An adjusted MVIM was recalculated using $\widehat{\mathcal{AMI}}_j = \widehat{\mathcal{CI}}_j \dfrac{\mathbb{V}(X_j)}{\hat{\sigma}^2_\nu}$. The adjusted MVIM also estimates the true MVIM for $X_1$ as $X_1$ has a linear relationship with the outcome. 
	\end{enumerate}
	
	The following three sets of Box plots shows the distribution of these three estimates,
	
	\begin{figure}[H]
		\centering
		\includegraphics[scale = 0.45]{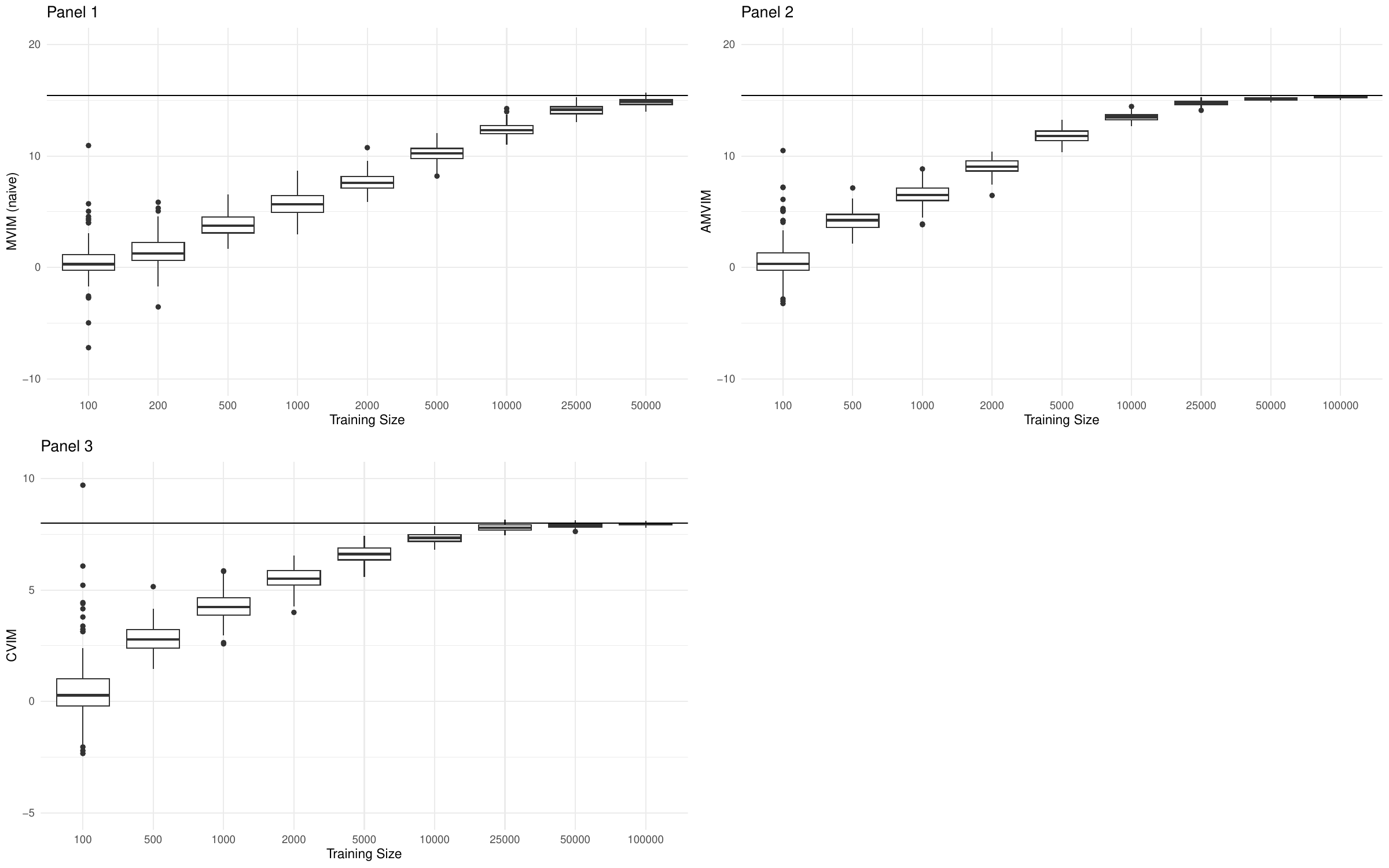}
		\caption{The distributions of the VIMs for weak dependence scenario. The first panel shows the distribution of the naive MVIM estimates, the second panel shows the distributions of the AVIMs obtained using equation \eqref{cgvim.lin.mod} and the third panel shows the distributions of the CVIMs}
		\label{Fig4.7}
	\end{figure}
	
	In this plot we can observe that all the estimates of MVIM and CVIM are converging to the truth. Here the true marginal MVIM was 15.45 and the true CVIM was 8.00. The AMVIM estimates (panel 2), were converging to the true value of MVIM at a faster rate compared to the naive MVIM estimates (panel 1).

	\begin{figure}[H]
		\centering
		\includegraphics[scale = 0.45]{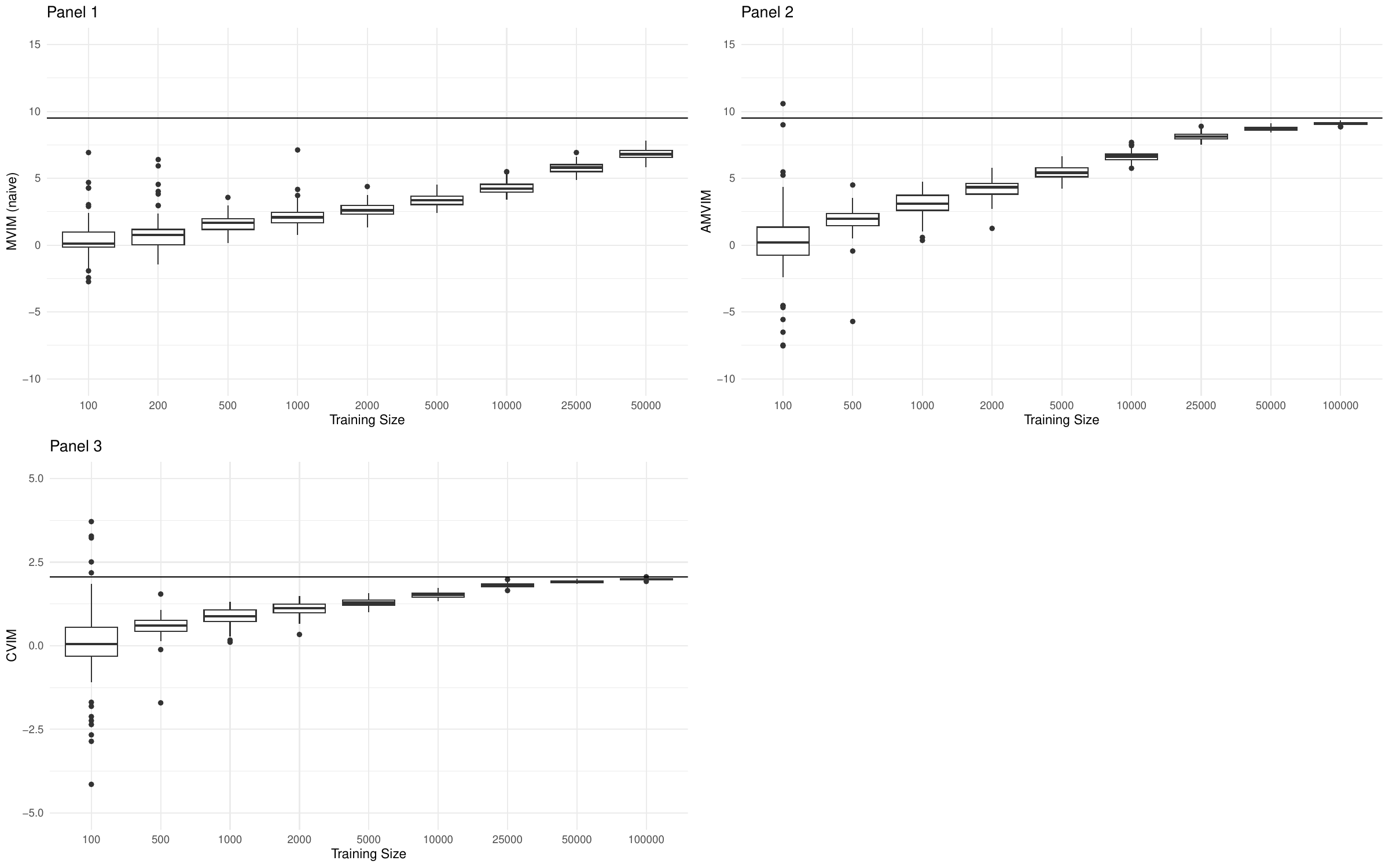}
		\caption{The distributions of the VIMs for moderate dependence scenario. The first panel shows the distribution of the naive MVIM estimates, the second panel shows the distribution of the AMVIMs obtained using equation \eqref{cgvim.lin.mod} and the third panel shows the distribution of the CVIMs}
		\label{Fig4.8}
	\end{figure}
	
	Figure \ref{Fig4.8} shows that both the AMVIM and the CVIM were converging to their corresponding true values with increasing training size.  The naive estimates of MVIM were also converging, but the convergence was at a slower rate and even for a large training size of 50000, the estimates still did not converge. Here the true MVIM was 9.45 and the true CVIM was 2.06.

	\begin{figure}[H]
		\centering
		\includegraphics[scale = 0.45]{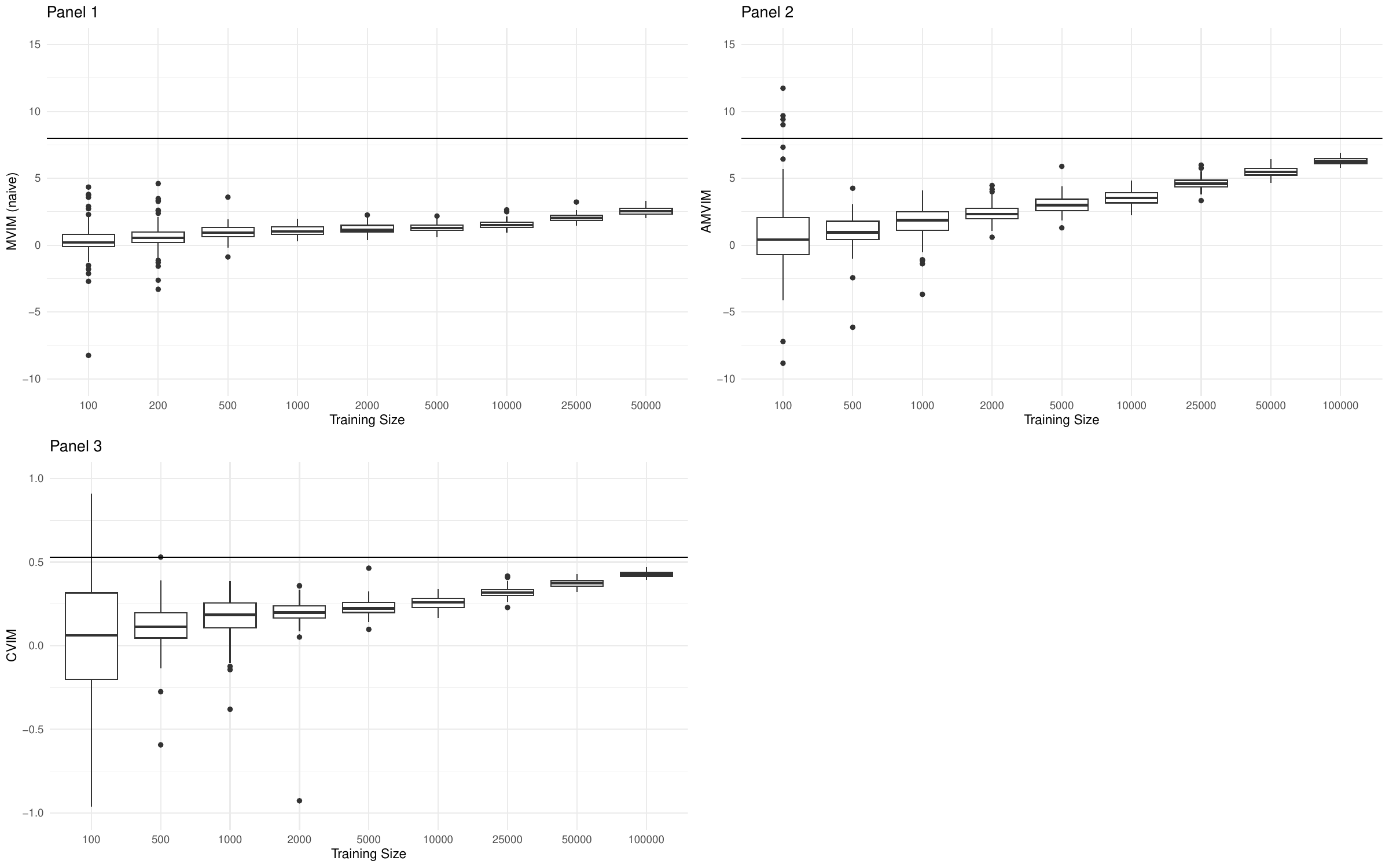}
		\caption{The distributions of the VIMs for strong dependence scenario. The first panel shows the distribution of the naive MVIM estimates, the second panel shows the distribution of the AMVIMs obtained using equation \eqref{cgvim.lin.mod} and the third panel shows the distribution of the CVIMs}
		\label{Fig4.9}
	\end{figure}
	
	Here the true MVIM here was 8.00 and the true CVIM was 0.53. All the estimates have a slower rate of convergence for strong dependence scenario compared to the previous scenarios. For training size 50000 none of the estimates converged to the true values.  However, the CVIM and the AMVIM estimates were still more accurate then the original MVIM estimates. This scenario represents a near violation of the positivity assumption and the lower value for CVIM compared to the AMVIM can address that. 
	
	\section{Discussion}
	
	In this research we proposed a population definition for a marginal variable importance metric and a conditional variable importance metric. The simulation results showed that when estimating from black box model such as XGBoost, the MVIM estimates were biased in the presence of correlation between the predictors even for a sufficiently large training size. This correlation distortion had also been mentioned by several authors, including \citet{verdinelli2024feature, strobl2007bias, strobl2008conditional, debeer2020conditional, hooker2021unrestricted, fisher2019all}, for MVIM type measures. When estimated from parametric models with correctly specified conditional expectation functions, the MVIM estimates were unbiased and consistent. For all our candidate models we investigated three distinct scenarios to evaluate the source of bias. In the completely independent scenario, the bias was only observed for small training sizes and reduced to zero with increasing training size. When two predictors were highly correlated, the bias in estimating MVIM increased compared to the completely independent scenario. In the multivariate correlation scenario, the bias in estimating MVIM did not converge to zero even for a training size of 50,000. The correlation distortion occurred due to the lack of extrapolation ability in black-box models such as XGBoost. \citet{verdinelli2024feature} discussed the bias due to correlation distortion in LOCO estimation as an inherent problem of black-box models like Random Forest. The second aim of this research was to investigate the cause of bias in estimating MVIM using bias-variance decomposition. We investigated the source of this bias using the bias-variance decomposition of both the switched and original prediction errors and derived the additional term $\delta_j$ from the decomposition. This term was the major reason for the first-order bias in MVIM estimates, as observed from Figures \ref{Fig4.1} to \ref{Fig4.3}. The additional bias term $\delta_j$, derived from the true and switched conditional expectations, was hard to estimate since obtaining an accurate estimation of the switched error from XGBoost was difficult in the presence of correlation distortion. Even if a prediction model like XGBoost produced nearly unbiased estimations for $f_0(\boldsymbol{X})$ for large training sizes, it still failed to produce accurate estimations for the switched conditional expectation $f_0(\boldsymbol{X}_{j^{\prime}})$ due to its lack of extrapolation ability in low-density regions in the sample space, which were induced by switching the predictor of interest. Thus, a highly accurate prediction model did not ensure accurate estimation of MVIM. In terms of a causal framework, MVIM estimates were biased in the case of a near violation of positivity assumptions. To our knowledge, this was the first instance where the bias-variance decomposition and its components had been derived and investigated for permutation-based importance measures. \\
	
	The next objective of this research was to address the bias induced by correlation distortion. Estimating the marginal importance of a predictor is challenging when predictors are correlated. Therefore, we focused on defining conditional importance based on the true conditional expectation function. Conditional permutation-based importance for Random Forests was proposed by \citet{strobl2008conditional}. Later, \citet{fisher2019all}, \citet{debeer2020conditional}, \citet{hooker2021unrestricted}, and \citet{chamma2024statistically} derived more general estimation techniques for conditional importance.The model-agnostic definition for conditional VIM (CVIM) was based on conditional switching, where predictors are switched while other predictors are fixed for two independent replicates from the population. We discussed the association of CVIM with CATE, a novel finding from this research. To estimate CVIM, a prediction model must be fit for each predictor of interest. In the presence of high correlation between predictors, CVIM is much smaller than MVIM; however, they are equal when predictors were independent. CVIM can also be obtained from causal variance decomposition as shown in \citet{chen2020causal}. If the estimated value of CVIM is close to zero, it can indicate either the variable is not conditionally important or a near violation of the positivity assumption. The near violation of the positivity assumption can be detected when the corresponding estimate of the AMVIM is much larger compared to the MVIM estimate. Additionally, CVIM is a quadratic function of CATE and can be used in the presence of treatment heterogeneity.   \\
	
	The bias variance decomposition was also derived for CVIM. The components of this decomposition were similar to the bias variance components obtained from the bias variance decomposition of MVIM. Similar to the decomposition of MVIM the CVIM decomposition also had an additional bias term $\delta_j$ for predictor $X_j$ with a different order of expectation as compared to the additional bias term obtained from MVIM decomposition. From our simulation results presented in Tables \ref{tab.cgvim.1} to \ref{tab.cgvim.3} and in Figures \ref{Fig4.4} to \ref{Fig4.6}, we observed that $\delta_j$ converges to zero with increasing training set size allowing the CVIM estimate to converge to the true value of CVIM. The convergence rate slows down with increasing correlation indicating a near violation of the positivity assumption. The convergence rate of CVIM estimate was faster when compared to the convergence rate of MVIM estimates, implying that CVIM estimates were less affected by the correlation distortion compared to the MVIM estimates. \\

	It is important to note that CVIM is measure for conditional importance, where MVIM is a measure of marginal importance of a predictor. As they are different measures, researchers need to identify which importance they are interested in. In presence of high correlation, CVIM estimates are more accurate than MVIM estimates and thus CVIM should be a preferred option. However, researcher might also be interested in marginal importance of a predictor. To address this issue we developed the adjusted MVIM which is a method to extract MVIM from CVIM. As can be seen from equation \eqref{cgvim.lin.mod} the AMVIM was obtained by multiplying the variance ratio $\frac{\mathbb{V}(X_j)} {\mathbb{V}(X_j\mid \boldsymbol{X}_{-j}) }$ of the predictor of interest to its CVIM, however, this relationship was only confirmed for linear and additive models which assume that the conditional variance of the predictor given the rest of the covariates is constant. The observations from Figure \ref{Fig4.7} to \ref{Fig4.9} indicate that the estimated AMVIM converges to the true value of MVIM at a faster rate compared to naive MVIM estimates with increasing training set sizes. All the importance measures defined in this study are model-agnostic and have causal interpretations, which relate to notions 2 and 3 of the importance explained in \citet{zhao2019causal}. \\
	
	The importance metrics which were defined in this research along with their estimation techniques have several key advantages over previously developed methods for explaining outputs from machine learning models. To our knowledge, this is the first time causal inference has been considered for both marginal and conditional importance directly derived from quadratic loss functions used in machine learning models. The importance metrics also enable the detection of any violation of the positivity assumption, which is a significant contribution of this research. Previous studies \citep{hooker2021unrestricted, strobl2008conditional, debeer2020conditional} have mentioned that in the presence of correlation distortion, the focus should be on estimating conditional importance rather than marginal importance. The AMVIM (which also estimate marginal importance) and its estimate proposed in this research are also novel, and it was demonstrated that they are less affected by correlation distortion compared to the nave MVIM estimates. Thus, in the presence of correlation distortion, researchers can rely on AMVIM to estimate the marginal importance of a predictor, which has been challenging to obtain, as noted by several other studies \citep{hooker2021unrestricted, strobl2008conditional, debeer2020conditional}. \\
	
	The estimation techniques for CVIM and adjusted AMVIM have some limitations. First of all, similar to the MVIM estimates for small sample sizes, the CVIM estimates can be inaccurate, specifically when the true prediction function is nonlinear and non-additive. This is again due to the extrapolation bias of XGBoost. However, the bias reduces to zero with larger training sets consistently. Furthermore, to estimate CVIM, one needs to fit a prediction model for each of the predictors. This can be a very intensive computational task given a large number of predictors. The process works fine when only a few predictors are of interest, as in causal inference. Computing standard errors and confidence intervals via Bootstrap adds to the computational expense. To estimate the AMVIM, this research only focused on the linear association between predictor and response. When the predictor and the response have a nonlinear and non-additive association, the AMVIM estimates could not be as easily extracted. \\

	Explaining outputs from machine learning methods has been an unresolved topic in the literature. Most of the proposed methods are data-driven and do not have any defined statistical parameter. These methods relied on complex machine learning models, with the relationships being explained through post-hoc analysis \citep{lundberg2017unified, ribeiro2016should}. \citet{rudin2019stop} argued that the common practice of using complex, opaque models and subsequently explaining them with post-hoc analysis was flawed and potentially dangerous. \citet{rudin2019stop} advocated for the direct use of inherently interpretable models, which could provide clear, understandable, and trustworthy insights without the need for separate explanation methods. This approach enhanced transparency, accountability, and fairness, making it more suitable for critical decision-making contexts where understanding the reasoning behind predictions was crucial. The MVIM, CVIM and AMVIM aimed to achieve those goals. By defining these importance measures, we extracted a statistical parameter directly from the prediction error/loss function of a model, making any machine learning model interpretable. Traditionally, permutation-based importance metrics were used for ranking predictors based on their importance in predicting the response variable and for feature selection. The importance measures described in this work could be estimated for both ranking predictors based on their contribution towards prediction and understanding their causal relationship with the outcome. \\
	
	Our future research will focus on refining methods to mitigate the impact of correlation distortion on MVIM and CVIM, particularly in scenarios with small sample sizes or highly nonlinear prediction functions. Enhancing the accuracy and computational efficiency of these methods will improve their robustness and applicability. Additionally, investigating non-linear associations between predictors and their effects on importance measures is crucial for capturing complex real-world data relationships. Future studies should also explore deriving importance measures from alternative objective functions beyond mean squared error, broadening the applicability of MVIM, CVIM, and AMVIM to various data types. Developing methods to determine the direction of association between predictors and outcomes, possibly through advanced visualization techniques or integrating causal inference frameworks, may further enhance the interpretability and practical utility of these measures. \\

\bibliographystyle{apalike}
\bibliography{kak_thesis}

\begin{thebibliography}{}

\bibitem[Austin, 2019]{austin2019assessing}
Austin, P.~C. (2019).
\newblock Assessing the performance of the generalized propensity score for
  estimating the effect of quantitative or continuous exposures on survival or
  time-to-event outcomes.
\newblock {\em Statistical methods in medical research}, 28(8):2348--2367.

\bibitem[Breiman, 1996]{breiman1996bagging}
Breiman, L. (1996).
\newblock Bagging predictors.
\newblock {\em Machine learning}, 24:123--140.

\bibitem[Breiman, 2001]{breiman2001random}
Breiman, L. (2001).
\newblock Random forests.
\newblock {\em Machine learning}, 45(1):5--32.

\bibitem[Buhlmann et~al., 2002]{buhlmann2002analyzing}
Buhlmann, P., Yu, B., et~al. (2002).
\newblock Analyzing bagging.
\newblock {\em Annals of statistics}, 30(4):927--961.

\bibitem[Chamma et~al., 2024]{chamma2024statistically}
Chamma, A., Engemann, D.~A., and Thirion, B. (2024).
\newblock Statistically valid variable importance assessment through
  conditional permutations.
\newblock {\em Advances in Neural Information Processing Systems}, 36.

\bibitem[Chen et~al., 2020]{chen2020causal}
Chen, B., Lawson, K.~A., Finelli, A., and Saarela, O. (2020).
\newblock Causal variance decompositions for institutional comparisons in
  healthcare.
\newblock {\em Statistical methods in medical research}, 29(7):1972--1986.

\bibitem[Chen and Guestrin, 2016]{chen2016xgboost}
Chen, T. and Guestrin, C. (2016).
\newblock Xgboost: A scalable tree boosting system.
\newblock In {\em Proceedings of the 22nd acm sigkdd international conference
  on knowledge discovery and data mining}, pages 785--794.

\bibitem[Cruz and Wishart, 2006]{cruz2006applications}
Cruz, J.~A. and Wishart, D.~S. (2006).
\newblock Applications of machine learning in cancer prediction and prognosis.
\newblock {\em Cancer informatics}, 2:117693510600200030.

\bibitem[Debeer and Strobl, 2020]{debeer2020conditional}
Debeer, D. and Strobl, C. (2020).
\newblock Conditional permutation importance revisited.
\newblock {\em BMC bioinformatics}, 21:1--30.

\bibitem[Diaz et~al., 2015]{diaz2015variable}
Diaz, I., Hubbard, A., Decker, A., and Cohen, M. (2015).
\newblock Variable importance and prediction methods for longitudinal problems
  with missing variables.
\newblock {\em PloS one}, 10(3).

\bibitem[Efron and Tibshirani, 1997]{efron1997improvements}
Efron, B. and Tibshirani, R. (1997).
\newblock Improvements on cross-validation: the 632+ bootstrap method.
\newblock {\em Journal of the American Statistical Association},
  92(438):548--560.

\bibitem[Fel et~al., 2021]{fel2021look}
Fel, T., Cad{\`e}ne, R., Chalvidal, M., Cord, M., Vigouroux, D., and Serre, T.
  (2021).
\newblock Look at the variance! efficient black-box explanations with
  sobol-based sensitivity analysis.
\newblock {\em Advances in neural information processing systems},
  34:26005--26014.

\bibitem[Fisher et~al., 2019a]{fisher2019all}
Fisher, A., Rudin, C., and Dominici, F. (2019a).
\newblock All models are wrong, but many are useful: Learning a variable's
  importance by studying an entire class of prediction models simultaneously.
\newblock {\em J. Mach. Learn. Res.}, 20(177):1--81.

\bibitem[Fisher et~al., 2019b]{fisher2019machine}
Fisher, C.~K., Smith, A.~M., and Walsh, J.~R. (2019b).
\newblock Machine learning for comprehensive forecasting of alzheimer’s
  disease progression.
\newblock {\em Scientific reports}, 9(1):1--14.

\bibitem[Friedman, 1991]{friedman1991multivariate}
Friedman, J.~H. (1991).
\newblock Multivariate adaptive regression splines.
\newblock {\em The annals of statistics}, 19(1):1--67.

\bibitem[Gregorutti et~al., 2017]{gregorutti2017correlation}
Gregorutti, B., Michel, B., and Saint-Pierre, P. (2017).
\newblock Correlation and variable importance in random forests.
\newblock {\em Statistics and Computing}, 27(3):659--678.

\bibitem[Hawken et~al., 2010]{hawken2010utility}
Hawken, S.~J., Greenwood, C.~M., Hudson, T.~J., Kustra, R., McLaughlin, J.,
  Yang, Q., Zanke, B.~W., and Little, J. (2010).
\newblock The utility and predictive value of combinations of low penetrance
  genes for screening and risk prediction of colorectal cancer.
\newblock {\em Human genetics}, 128(1):89--101.

\bibitem[Hooker, 2007]{hooker2007generalized}
Hooker, G. (2007).
\newblock Generalized functional anova diagnostics for high-dimensional
  functions of dependent variables.
\newblock {\em Journal of computational and graphical statistics},
  16(3):709--732.

\bibitem[Hooker et~al., 2021]{hooker2021unrestricted}
Hooker, G., Mentch, L., and Zhou, S. (2021).
\newblock Unrestricted permutation forces extrapolation: variable importance
  requires at least one more model, or there is no free variable importance.
\newblock {\em Statistics and Computing}, 31:1--16.

\bibitem[Howell et~al., 2012]{howell2012symptom}
Howell, D., Husain, A., Seow, H., Liu, Y., Kustra, R., Atzema, C., Dudgeon, D.,
  Earle, C., Sussman, J., and Barbera, L. (2012).
\newblock Symptom clusters in a population-based ambulatory cancer cohort
  validated using bootstrap methods.
\newblock {\em European Journal of Cancer}, 48(16):3073--3081.

\bibitem[Kaneko, 2022]{kaneko2022cross}
Kaneko, H. (2022).
\newblock Cross-validated permutation feature importance considering
  correlation between features.
\newblock {\em Analytical Science Advances}, 3(9-10):278--287.

\bibitem[Lei et~al., 2018]{lei2018distribution}
Lei, J., G’Sell, M., Rinaldo, A., Tibshirani, R.~J., and Wasserman, L.
  (2018).
\newblock Distribution-free predictive inference for regression.
\newblock {\em Journal of the American Statistical Association},
  113(523):1094--1111.

\bibitem[Libbrecht and Noble, 2015]{libbrecht2015machine}
Libbrecht, M.~W. and Noble, W.~S. (2015).
\newblock Machine learning applications in genetics and genomics.
\newblock {\em Nature Reviews Genetics}, 16(6):321--332.

\bibitem[Liestbl et~al., 1994]{liestbl1994survival}
Liestbl, K., Andersen, P.~K., and Andersen, U. (1994).
\newblock Survival analysis and neural nets.
\newblock {\em Statistics in medicine}, 13(12):1189--1200.

\bibitem[Lundberg and Lee, 2017]{lundberg2017unified}
Lundberg, S.~M. and Lee, S.-I. (2017).
\newblock A unified approach to interpreting model predictions.
\newblock In {\em Proceedings of the 31st international conference on neural
  information processing systems}, pages 4768--4777.

\bibitem[Molnar, 2020]{molnar2020interpretable}
Molnar, C. (2020).
\newblock {\em Interpretable machine learning}.
\newblock Lulu. com.

\bibitem[Petersen et~al., 2012]{petersen2012diagnosing}
Petersen, M.~L., Porter, K.~E., Gruber, S., Wang, Y., and Van Der~Laan, M.~J.
  (2012).
\newblock Diagnosing and responding to violations in the positivity assumption.
\newblock {\em Statistical methods in medical research}, 21(1):31--54.

\bibitem[Ribeiro et~al., 2016]{ribeiro2016should}
Ribeiro, M.~T., Singh, S., and Guestrin, C. (2016).
\newblock " why should i trust you?" explaining the predictions of any
  classifier.
\newblock In {\em Proceedings of the 22nd ACM SIGKDD international conference
  on knowledge discovery and data mining}, pages 1135--1144.

\bibitem[Rinaldo et~al., 2019]{rinaldo2019bootstrapping}
Rinaldo, A., Wasserman, L., and G’Sell, M. (2019).
\newblock Bootstrapping and sample splitting for high-dimensional,
  assumption-lean inference.
\newblock {\em The Annals of Statistics}, 47(6):3438--3469.

\bibitem[Ripley and Ripley, 2001]{ripley2001neural}
Ripley, B.~D. and Ripley, R.~M. (2001).
\newblock Neural networks as statistical methods in survival analysis.
\newblock {\em Clinical applications of artificial neural networks}, pages
  237--255.

\bibitem[Rudin, 2019]{rudin2019stop}
Rudin, C. (2019).
\newblock Stop explaining black box machine learning models for high stakes
  decisions and use interpretable models instead.
\newblock {\em Nature machine intelligence}, 1(5):206--215.

\bibitem[Strobl et~al., 2008]{strobl2008conditional}
Strobl, C., Boulesteix, A.-L., Kneib, T., Augustin, T., and Zeileis, A. (2008).
\newblock Conditional variable importance for random forests.
\newblock {\em BMC bioinformatics}, 9:1--11.

\bibitem[Strobl et~al., 2007]{strobl2007bias}
Strobl, C., Boulesteix, A.-L., Zeileis, A., and Hothorn, T. (2007).
\newblock Bias in random forest variable importance measures: Illustrations,
  sources and a solution.
\newblock {\em BMC bioinformatics}, 8(1):1--21.

\bibitem[Tibshirani and Efron, 1993]{tibshirani1993introduction}
Tibshirani, R.~J. and Efron, B. (1993).
\newblock An introduction to the bootstrap.
\newblock {\em Monographs on statistics and applied probability}, 57:1--436.

\bibitem[Van~der Laan and Rose, 2011]{van2011targeted}
Van~der Laan, M.~J. and Rose, S. (2011).
\newblock {\em Targeted learning: causal inference for observational and
  experimental data}.
\newblock Springer Science \& Business Media.

\bibitem[Verdinelli and Wasserman, 2024]{verdinelli2024feature}
Verdinelli, I. and Wasserman, L. (2024).
\newblock Feature importance: A closer look at shapley values and loco.
\newblock {\em Statistical Science}, 39(4):623--636.

\bibitem[Weller et~al., 2021]{weller2021predicting}
Weller, O., Sagers, L., Hanson, C., Barnes, M., Snell, Q., and Tass, E.~S.
  (2021).
\newblock Predicting suicidal thoughts and behavior among adolescents using the
  risk and protective factor framework: A large-scale machine learning
  approach.
\newblock {\em Plos one}, 16(11):e0258535.

\bibitem[Wood, 2017]{mgcv}
Wood, S. (2017).
\newblock {\em Generalized Additive Models: An Introduction with R}.
\newblock Chapman and Hall/CRC, 2 edition.

\bibitem[Zhao and Hastie, 2019]{zhao2019causal}
Zhao, Q. and Hastie, T. (2019).
\newblock Causal interpretations of black-box models.
\newblock {\em Journal of Business \& Economic Statistics}, pages 1--10.

\end{thebibliography}

\end{document}